%% file: 0-main-tacl.tex
\documentclass[11pt,a4paper]{article}
\usepackage{times,latexsym}
\usepackage{url}
\usepackage[T1]{fontenc}
\usepackage{ragged2e}
\usepackage{booktabs}
\usepackage{multirow}
\usepackage{tabularx}
\usepackage{float}
\usepackage[acceptedWithA]{tacl2021v1}
\usepackage{tacl2021v1}

\usepackage{xspace,mfirstuc,tabulary}

\newif\iftaclinstructions
\taclinstructionsfalse 
\iftaclinstructions
\renewcommand{\confidential}{}
\renewcommand{\anonsubtext}{(No author info supplied here, for consistency with
TACL-submission anonymization requirements)}
\newcommand{\instr}
\fi

\iftaclpubformat 

\else

\fi

\usepackage{booktabs}
\usepackage{listings}
\usepackage{enumitem}
\usepackage{xcolor}
\usepackage{multirow}
\usepackage{graphicx}
\usepackage{subcaption}
\usepackage[most]{tcolorbox} 
\usepackage{listings}
\usepackage{listings}

\usepackage{relsize,etoolbox}
\AtBeginEnvironment{quote}{\smaller}

\definecolor{mycolor_1}{HTML}{FFC000}
\definecolor{mycolor_2}{HTML}{D9E1F2}

\renewcommand{\lstlistingname}{Example}
\lstdefinestyle{datalogstyle}{
    basicstyle=\color{darktext}\ttfamily\fontsize{7}{8}\selectfont,  %
	xleftmargin={1pt},
    xrightmargin={1pt},
    columns=flexible,
    breakindent=0pt,
    breaklines=true, 
    frame=tb,
    stepnumber=1,
    firstnumber=1,
    numberfirstline=true,
    tabsize=2,
    extendedchars=true,
    breaklines=true,
    columns=fullflexible,
    keepspaces=true,
    escapeinside={@}{@},
    firstnumber=last,
    captionpos=b,
	commentstyle=\color{black!65},
	numberstyle=\tiny\color{black!65},
    stringstyle=\color{codepurple},
    breakatwhitespace=false, 
    keepspaces=true,              
    mathescape=true, 
    numbersep=5pt,                  
    showspaces=false,                
    showstringspaces=false,
    showtabs=false,
    aboveskip={0.2\baselineskip},
    belowskip={0.2\baselineskip},
}
\definecolor{mygray}{RGB}{169,169,169}
\definecolor{darktext}{RGB}{0,0,0}
\definecolor{darkblue}{RGB}{0,0,139}

\title{In Generative AI We (Dis)Trust? \\Computational Analysis of Trust and Distrust in Reddit Discussions}

\author{ \bf
Aria Pessianzadeh, 
Naima Sultana, 
Hildegarde Van den Bulck, \\
\bf
David Gefen, 
Shahin Jabbari, 
Rezvaneh Rezapour \\
Drexel University \\ 
\texttt{\{ap3943, shahin, shadi.rezapour\}@drexel.edu}
}

\date{}

\begin{document}
\maketitle


\input{01-abstract}

\input{02-intro}
\input{03-relatedwork}
\input{04-method}

\input{05-results}

\input{06-discussion}

\input{08-ethics_limitation}
\input{10-acknowledgement}

\bibliography{custom}
\bibliographystyle{acl_natbib}



\appendix
\input{09-appendix}
\end{document}

%% file: 01-abstract.tex
\begin{abstract}
The rise of generative AI (GenAI) has impacted many aspects of human life. As these systems become embedded in everyday practices, understanding public trust in them also becomes essential for responsible adoption and governance. Prior work on trust in AI has largely drawn from psychology and human–computer interaction, but there is a lack of computational, large-scale, and longitudinal approaches to measuring trust and distrust in GenAI and large language models (LLMs).
This paper presents the first computational study of \textit{Trust} and \textit{Distrust} in GenAI, using a multi-year Reddit dataset (2022–2025) spanning 39 subreddits and 230,576 posts. Crowd-sourced annotations of a representative sample were combined with classification models to scale analysis. We find that \textit{Trust} and \textit{Distrust} are nearly balanced over time, although \textit{Trust} modestly outweighs \textit{Distrust} with shifts around major model releases. Technical performance and usability dominate as dimensions, while personal experience is the most frequent reason shaping attitudes. Distinct patterns also emerge across trustors (e.g., experts, ethicists, general users). Our results provide a methodological framework for large-scale \textit{Trust} analysis and insights into evolving public perceptions of GenAI.

\end{abstract}\label{sec:abstract}

%% file: 02-intro.tex
\section{Introduction}\label{sec:intro} 

Generative artificial intelligence (GenAI) and large language models (LLMs) are transforming how people search for information, create content, and interact online~\cite{ferrara2024genai,liang2025widespread}. Their rapid adoption in classrooms and workplaces~\cite{ghimire2024generative,pettersson2024generative}, as well as in creative industries~\cite{amato2019ai}, has triggered global debates about the potential benefits and risks of these technologies~\cite{bender2021dangers,crawford2021atlas}. At the center of these debates lies the question of trust: will the public perceive GenAI as reliable, competent, and aligned with their values, or as opaque, biased, and potentially harmful? 

Trust is widely recognized as a prerequisite for adoption, appropriate use, and governance of AI systems \cite{mayer1995integrative,hancock2020ai,lee2004trust,hoff2015trust}.
Public attention to GenAI, and especially to questions of trust and reliability, has surged in parallel with the release of increasingly advanced LLMs. The launch of ChatGPT in late 2022, followed by successive breakthroughs such as GPT-4 and LLaMA2, amplified both enthusiasm and concern, making trust a focal point of GenAI discourse \cite{shen2023chatgpt,weidinger2022taxonomy}. Understanding how trust and distrust are expressed in response to such advancements is crucial for capturing the dynamics of public attitudes and informing the responsible design and governance of GenAI. 

Prior research on trust in GenAI has mainly emphasized theoretical frameworks or small-scale qualitative studies \cite{schuetz2024qualitative, balayn2024empirical}. Recent computational work has analyzed public sentiment toward LLMs using approaches such as Critical Discourse Analysis \cite{Heaton2024ChatGPT, Emdad2023ChatGPTAF}, but focused on domain-specific contexts like education or health care rather than systematically mapping diverse dimensions of trust and distrust. Attitudes have also been examined through validated scales such as AT-GLLM and AT-PLLM \cite{liebherrdeveloping}, while TAM-based studies showed that trust, privacy, and misuse concerns strongly shape AI adoption \cite{abdaljaleel2023factors, kelly2025factors}. Although prior work has developed valuable conceptual frameworks for trust in GenAI and LLMs, there is limited large-scale, longitudinal, computational analysis of how public attitudes evolve around GenAI. 

We address this gap with a multi-year analysis of Reddit discussions (Nov 2022–Jun 2025). Reddit offers a suitable corpus given its scale, topical diversity, and sustained discourse around model releases. We collected 230,576 GenAI-related posts from 39 subreddits and crowd-sourced annotations on a representative sample to capture \textit{Trust}, \textit{Distrust}, their dimensions, reasons, and trustors, then scaled these analyses using transformer-based and state-of-the-art LLMs. Our findings show that \textit{Trust} moderately outweighs \textit{Distrust} across the timeline, with shifts around major releases. At the dimension level, \textit{Competence}, \textit{Reliability}, and \textit{Familiarity} dominate, underscoring functionality and personal use as key drivers. Personal experience emerges as the most salient reason for (dis)trust, consistent with recent industry reports~\cite{openai2025_how_people_using_chatgpt,anthropic_claude_support_advice_companionship}. Finally, different trustors (e.g., experts, ethicists, business leaders) exhibit distinct patterns, highlighting the sociotechnical complexity of GenAI discourse.

This work contributes the first large-scale dataset of Reddit posts on GenAI labeled for \textit{Trust}, \textit{Distrust}, dimensions, reasons, and trustors; a methodological framework for classifying \textit{Trust} and \textit{Distrust} in social media discourse; and offers new empirical insights into how \textit{Trust} and \textit{Distrust} toward GenAI evolve, what dimensions dominate, and how different actors express them. By combining computational analysis with sociotechnical framing, our study advances the understanding of trust in GenAI and provides tools for monitoring and contextualizing public perceptions. The de-identified data and codes will be released upon paper's accepntance.  

%% file: 03-relatedwork.tex
\section{Related Work} \label{sec:relatedwork}

\noindent\textbf{Trust in Technology \& AI Systems. }
Foundational research on trust in technology provides key perspectives on how trust is formed. Trust in the social sciences is defined as the willingness to be vulnerable by relying on another, setting aside concerns about potential negative behavior~\cite{mayer1995integrative,luhmann2018trust}. \citet{schuetz2024qualitative} found perceptions and experiences with technology, trust transfer, and trust calculus as three categories of influential factors that shape trust. Similarly, \citet{lee2004trust} highlighted how context, system characteristics, and cognitive processes influence trust relationships between users and automated systems. \citet{glikson2020human} show that cognitive trust stems from transparency and reliability, while emotional trust arises through anthropomorphism. \citet{jacovi2021formalizing} formalized AI–human trust as a cognitive model grounded in user vulnerability and anticipation of AI decisions, where trust depends on expectations that agreements will hold.
\citet{balayn2024empirical} emphasize the complexity of LLM supply chains, where trust dynamics extend from foundational model developers to downstream companies and end users, influenced by both personal propensity and context of use. In parallel, \citet{kim2023humans} point to the interplay of human factors (e.g., domain knowledge), AI factors (e.g., ability, integrity, benevolence), and contextual conditions (e.g., task difficulty, perceived risks) in determining levels of trust in AI systems.
These mechanisms also apply in domains such as electronic marketplaces \cite{pavlou2004building}. Importantly, distrust is not the absence of trust but distinct expectations of negative behavior~\cite{lewicki1998trust}, supported by evidence that trust and distrust rely on different neuro-cognitive processes~\cite{dimoka2010does}. Trust generally increases behavioral intentions, whereas distrust reduces them~\cite{gefen2025importance}.

\noindent\textbf{Computational Analysis of Trust. } 
Recent work has developed computational and survey-based measures of trust. The Attitudes Toward General LLMs (AT-GLLM) and Attitudes Toward Primary LLMs (AT-PLLM) scales were recently developed and validated as reliable tools for assessing acceptance and fear in LLM contexts, distinguishing between societal perceptions and individual user experiences \cite{liebherrdeveloping}. Computational analyses of public discourse also revealed mixed sentiment: topic and sentiment modeling of Reddit discussions about ChatGPT and DeepSeek found recurring concerns around bias, ethical risks, and transparency \cite{katta2025analyzing}. Building on the Technology Acceptance Model (TAM), \citet{abdaljaleel2023factors} introduced the TAME-ChatGPT instrument to study healthcare students' attitudes, while \citet{kelly2025factors} found that trust and privacy concerns differentially influence ChatGPT acceptance in mental versus physical healthcare. 

\noindent\textbf{LLMs and Trust. }
Research specific to LLMs highlights the multi-layered nature of trust and distrust.
\citet{paraschou2025ties} proposed the `bowtie model' to conceptualize trustor–trustee relationships, showing how perceptions of competence, knowledge, and transparency shape user behavior. \citet{davoodi2024large} applied computational methods to customer feedback and identified delivery processes as central to trust formation. Broader reviews highlight conceptual fragmentation: \citet{mehrotra2024systematic} stressed definitional inconsistencies, while \citet{sousa2023challenges} argued that trust is often oversimplified. 
\citet{gerlich2024exploring} found that many individuals trust AI more than humans, attributing this to their perceived impartiality, but other studies found otherwise~\cite{gefen2025importance}, suggesting that this might be context-dependent. 
Recent mapping studies confirm rising attention to LLM trustworthiness, with transparency, explainability, and reliability as recurring themes \cite{de2025mapping}.
Computational studies further analyze public opinion toward LLMs across social platforms. \citet{Heaton2024ChatGPT} show that Twitter users often frame ChatGPT as a social actor, while \citet{Emdad2023ChatGPTAF} and \citet{takagi-2023-banning} examine Reddit discussions on ChatGPT in education, finding mostly neutral or contested views. 
\citet{Skjuve2023User} emphasizes pragmatic and hedonic features in shaping trust. 
Other work uses machine learning for trust modeling \cite{zapata2023trust,yahyaoui2016feature,Wang2020Survey,Lopez2015Trust,Hauke2013supervised}, while \citet{gefen2024evolving} demonstrates that perceptions of trust in GenAI evolve from technical concerns toward broader societal issues.
Prior work provides valuable conceptual and computational insights but remains fragmented, domain-specific, and small in scale. Few studies track how trust and distrust in GenAI evolve in response to major model releases, a gap this study addresses.

%% file: 04-method.tex
\section{Method} \label{sec:method} 

\subsection{Data Collection}
To examine discussions surrounding GenAI on Reddit following the release of ChatGPT (November 30, 2022), we implemented a two-step data collection strategy. We first identified relevant subreddits and then applied content-based filtering to isolate GenAI-related posts.

\noindent\textbf{Step 1: Identification of Subreddits.}
We systematically identified communities focused on LLMs, their developers, artificial intelligence more broadly, and related technologies. We reviewed approximately 90 candidate subreddits, examining their content, ``About'' pages, topical focus, and creation dates. Based on relevance and sustained activity, we selected 39 subreddits that were active between November 30, 2022, and June 30, 2025. Using Pushshift~\cite{Baumgartner2020Pushshift} and Arcticshift~\cite{Heitmann2025arctic_shift}, we retrieved 1,944,899 posts from these subreddits.

\noindent\textbf{Step 2: Identification of GenAI-Related Posts.}
To identify posts specifically related to GenAI within this corpus, we employed a mixed-method approach combining keyword-based filtering and topic modeling. We first compiled a list of seed terms related to GenAI and expanded it to include the names of major models (e.g., prominent LLMs), as well as key concepts and topics associated with generative AI.
To further refine and expand this list, we applied Latent Dirichlet Allocation (LDA)~\cite{blei2009topic} to posts from the two highest-activity subreddits (\textit{r/ChatGPT} and \textit{r/OpenAI}) and two broader AI-focused subreddits (\textit{r/ArtificialIntelligence} and \textit{r/MachineLearning}). We examined the resulting topics to identify additional relevant terms and incorporated them into our keyword list. Using this comprehensive set of GenAI-related keywords, we retained posts containing at least one matching term (see Appendix~\ref{sec:app-data} for the list of subreddits/keywords).

We excluded posts in which the author or content was marked as ``removed'' or ``deleted'' (to protect user privacy),  posts with empty bodies, and posts containing fewer than five words. After applying these filtering criteria, the final analytic dataset comprised 230,575 posts.

\subsection{Conceptual Frameworks} 
\subsubsection{Dimensions of Trust and Distrust}
Trust and distrust are central but distinct constructs in human–AI interaction. Drawing from psychology, sociology, and HCI research, we define \textit{Trust} as the willingness to rely on a system with positive expectations of its reliability, competence, and intentions, even under conditions of uncertainty or vulnerability \cite{mayer1995integrative, rousseau1998not}. In contrast, \textit{Distrust} refers to active suspicion and negative expectations, grounded in concerns about unreliability, deception, or potential harm \cite{lewicki1998trust, harrison2001trust}. Importantly, \textit{Distrust} is not merely the absence of \textit{Trust}; rather, it is a qualitatively distinct phenomenon with its own logic and consequences \cite{lewicki1998trust}.

Grounded in extensive literature review of trust in technology and AI \cite{zhang2023we, mcknight2009trust, bach2024systematic, braga2018survey, xu2014different, lankton2015technology, zhang2024profiling, kim2024m, klingbeil2024trust, salimzadeh2024dealing, shin2021effects, blobaum2016trust, mcknight2017distinguishing, horowitz2024adopting, schuetz2025qualitative}, we first compiled 37 candidate dimensions of \textit{Trust} and \textit{Distrust} (see Appendix \ref{sec:app:dimensions-litearture}).
A team of four annotators manually coded a sample of 100 posts to evaluate the relevance and applicability of these dimensions in the context of GenAI. Through iterative rounds of coding and discussion, we distilled the list into 6 key dimensions of \textit{Trust} (\textit{Reliability, Competence, Familiarity, Integrity, Transparency, Benevolence}) and 7 key dimensions of \textit{Distrust} (\textit{Unreliability, Incompetence, Unfamiliarity, Deception, Dishonesty, Opaqueness, Malevolence}), adapted to LLMs and GenAI (see Table \ref{tab:trust_distrust_combined}). 

\begin{table}[t]
\centering
\resizebox{\columnwidth}{!}{%
\begin{tabular}{lp{2cm}p{8cm}}
\toprule
 & \textbf{Dimension} & \textbf{Definition} \\
\midrule 
 & Reliability  & Information accuracy and consistency\\
 & Transparency & Openness about GenAI processes and data use \\
\multirow{3}{*}{\rotatebox{90}{Trust}} & Familiarity  & Previous exposure to GenAI or similar systems \\
 & Integrity  & Adherence to ethical principles acceptable to users\\
 & Competence  & Effective functionality and features \\
 & Benevolence & Belief that GenAI acts with good intentions \\
\midrule  
 & Unreliability & Lack of accuracy and consistency \\
 & Opaqueness & Lack of transparency in data use \\
\multirow{4}{*}{\rotatebox{90}{Distrust}} & Unfamiliarity & New or unfamiliar technology causing doubt \\
 & Dishonesty & Failure to follow user-accepted ethical principles\\
 & Incompetence & Inability to perform tasks effectively \\
 & Deception & Perception of dishonesty or hallucination in outputs \\
 & Malevolence & Belief that GenAI has harmful or malicious intent\\
\bottomrule
\end{tabular}%
}
\caption{Dimensions of \textit{Trust} and \textit{Distrust}}
\label{tab:trust_distrust_combined}
\vspace{-0.5cm}
\end{table}

\subsubsection{Trustor and Reason of Trust} 
We further identified two additional concepts: (1) the \textit{trustor}, i.e., who expresses trust or distrust, and (2) the underlying \textit{reason} for that judgment.

\noindent\textbf{Trustor.}
In trust relationships, the \textit{trustor} is the entity placing confidence or suspicion in a system, while the \textit{trustee} is the system or actor being judged~\cite{mayer1995integrative, paraschou2025ties}. In GenAI, the trustor may range from end-users to institutions shaping AI narratives. Their identity influences both expectations and evaluation criteria~\cite{balayn2024empirical, kim2023humans}; e.g., developers may emphasize transparency and reliability, while ethicists focus on fairness and accountability. To capture this diversity, we defined 10 categories through inductive coding: GenAI users, software developers, researchers or academics, tech industry professionals, general public, media and journalists, business leaders or executives, AI ethicists or advocacy groups, artists or creatives, and educators or knowledge workers.

\noindent\textbf{Reason.}
The \textit{reason} for trust or distrust reflects the evidence upon which the trustor bases their judgment. Social psychology and HCI research show that reasons stem from prior experience, social influence, or perceived organizational responsibility~\cite{lee2004trust, glikson2020human}. In online discourse, these may be direct (e.g., personal use) or indirect (e.g., media or peer reports). We defined 6 categories: personal experience using AI models, reports from media or experts, discussions on social media or forums, general perceptions of AI technology, general perceptions of responsible companies, and experiences shared by friends, family, or colleagues.

\subsection{Crowdsourcing and Data Annotation}\label{sec:crowdsourcing}
We developed a codebook with definitions and examples of \textit{Trust} and \textit{Distrust} dimensions to guide the annotation process. We opted for crowdsourcing rather than expert-only annotation to leverage the broader demographic diversity and scalability it provides \cite{sap2021annotators, wan2023everyone}.
The annotation task was hosted on Potato \cite{pei2022potato} and deployed through \href{https://www.prolific.com/}{Prolific}. Annotators received the detailed guidelines and were tasked to select: (1) if a post expressed \textit{Trust}, \textit{Distrust}, \textit{Both}, or \textit{Neither} toward GenAI; (2) relevant dimensions of \textit{Trust} or \textit{Distrust}; (3) the \textit{trustor} (the entity expressing \textit{Trust} or \textit{Distrust}); and (4) underlying \textit{reason} from our predefined list. 
\begin{table}[ht!]
\centering
\resizebox{\columnwidth}{!}{%
\begin{tabular}{p{1.1cm}p{8cm}}
\toprule
\multirow{8}{*}{\textbf{Trust}} & I love having ChatGPT explain technical aspects to me as if I were a medieval noble receiving counsel from a trusted advisor. My life becomes so much less mundane. \\ \cline{2-2}
&Currently using ChatGPT to create a cheat sheet for units in my physics. This is 100 times easier to learn as well. Did I just make my teacher's job obsolete? \\ \midrule
 
\multirow{9}{*}{\textbf{Distrust}} & GPT4 is working really badly all of a sudden. Anyone else noticing this? It takes forever, and fails to respond with an error. \\ \cline{2-2}
& Why GPT4 just no longer follow instructions? Is it just for me, or is GPT4 no longer following instructions? Simple prompts that have always worked are now suddenly misinterpreted by ChatGPT. \\ \midrule
\multirow{18}{*}{\textbf{Both}} & The accuracy of the responses generated by ChatGPT depends on many factors, e.g., quality of the input and complexity of the task. Generally, ChatGPT has shown to produce highly coherent and human-like responses, especially for tasks that involve NLP, language translation, and generating textual content. However, like any AI model, ChatGPT can sometimes make errors or produce inaccurate responses, particularly where it lacks context or faces new or unexpected situations. \\\cline{2-2}
& What if most answers from ChatGPT are wrong? But we just oversaw, as we are still amazed by the awesomeness of quickly responding to complicated inquiries. \\\bottomrule
\end{tabular}}
\caption{Examples of posts labeled as \textit{Trust}, \textit{Distrust}, and \textit{Both}}
\label{tab:Examples for labels}
\vspace{-5mm}
\end{table}

Each post was independently annotated by five annotators to ensure labeling reliability. Across three annotation rounds, annotators labeled $2,690$ posts. For the first task, we only retained posts for which at least three out of five annotators agreed on the label (choosing the majority vote); posts without majority agreement were discarded. After applying quality control, $2,227$ posts were retained. Inter-annotator reliability was assessed using Fleiss' $\kappa$, resulting in $\kappa = 0.42$, with an average agreement rate of 77\%. Among the retained posts, 877 (39\%) had agreement from three out of five annotators, 800 (36\%) had agreement from four out of five annotators, and 550 (25\%) achieved unanimous agreement.
The codebook and annotation guidelines are in Appendix~\ref{sec:app-annot-process}. Table~\ref{tab:Examples for labels} presents two example posts for each class.

For dimensions of trust and distrust, we only used posts kept from the previous task. For each post, we retained the dimension annotations provided by the annotators whose labels formed the majority in the previous task. To construct ground-truth labels for the dimensions, we applied two aggregation schemes: \textit{Majority}: dimensions selected by more than half of the annotators, and \textit{Any}: dimensions selected by at least one of the annotators. 
To compute inter-annotator agreement for dimensions, we evaluated trust dimensions only for posts labeled \textit{Trust} and distrust dimensions only for posts labeled \textit{Distrust}. The average agreement across the 13 dimensions was $0.82$.
Similarly, for classifying Trustor and Reason, the final labels were obtained based on majority vote. Average agreement rate between annotators was $0.72$ for \textit{Trustor} and $0.81$ for \textit{Reason}.

\subsection{Classification}

\subsubsection{Trust and Distrust}  
We employed a multi-label classification framework with four labels in task 1: \textit{Trust}, \textit{Distrust}, \textit{Both}, and \textit{Neither}. Our annotated dataset was split into training, validation, and test sets using a 0.7:0.1:0.2 ratio, stratified by label to preserve class distribution. The splits contained $1,558$, $223$, and $446$ posts for train, validation, and test, respectively.  

\noindent\textbf{Transformer-based Models.}  
We fine-tuned five pretrained transformer models, including BERT \cite{DBLP:journals/corr/abs-1810-04805}, RoBERTa \cite{DBLP:journals/corr/abs-1907-11692}, DistilBERT \cite{sanh2019distilbert}, ModernBERT \cite{warner2024smarter}, and SocBERT \cite{guo2023socbert}, on the training set. Fine-tuning was conducted with the \texttt{AdamW} optimizer, \texttt{batch size} $8$, \texttt{learning rate} $2e-5$, and up to $3$ \texttt{epochs} with early stopping based on validation loss. Hyperparameters were selected through validation performance.  

\noindent\textbf{LLMs.}  
We evaluated eleven OpenAI models (GPT-5.2-2025-12-11, GPT-5.1-2025-11-13, GPT-5-2025-08-07, GPT-5-mini-2025-08-07, GPT-5-nano-2025-08-07, GPT-4o-2024-11-20, GPT-4o-mini-2024-07-18, GPT-4.1-2025-04-14, GPT-4.1-mini-2025-04-14, GPT-4.1-nano-2025-04-14, GPT-oss) \cite{openai-GPT41, openai-GPT41mini, openai-GPT4o, openai-GPT4omini, openai-GPToss, openai-GPT5mini, openai-GPT5nano, openai-GPT4turbo, openai-GPT5} and two open-source LLMs (Llama3-8B \cite{llama3modelcard}, and Mixtral-8x7B \cite{jiang2024mixtral}). For these models, we designed both zero-shot and few-shot prompts (see Appendix~\ref{sec:app-prompts}). Few-shot prompts included $3$ examples per label from the same validation set. We set \texttt{temperature} = $0.0$.

\noindent\textbf{Evaluation.}  
We evaluated all models on the held-out test set. Weighted F1, per-class precision, and recall were used as metrics.

\subsubsection{Dimensions of Trust and Distrust}
Building on the previous task, we extended the classification framework to predict the 13 fine-grained dimensions of \textit{Trust} and \textit{Distrust} defined in Table~\ref{tab:trust_distrust_combined}. Due to limited annotated data and the sparsity of several dimensions, we relied exclusively on LLMs for this task, as the dataset was insufficient to train or fine-tune supervised transformer-based classifiers.
For each post, the model was prompted to identify one or more applicable dimensions. To avoid excessively long prompts and to ensure category relevance, the applicable prompt was conditioned on the reported label: if the post was labeled as \textit{Trust} in Task 1, the model was prompted only with trust dimensions; if \textit{Distrust}, only distrust dimensions; and if \textit{Both}, with both sets of dimensions.  

\noindent\textbf{Prompting Setup.}  
We designed both zero-shot and few-shot prompts, including $3$ illustrative cases per dimension in the few-shot setting. All prompts included the definitions of dimensions from our codebook (see Appendix~\ref{sec:app-prompts}). We set \texttt{Temperature} = $0.0$ and \texttt{top-p} = $1.0$. For each post, the model could return multiple dimensions. 

\noindent\textbf{Evaluation.}  
We evaluate model performance under two groundtruth label aggregation schemes defined in \S~\ref{sec:crowdsourcing}: \textit{Majority} and \textit{Any}. Treating each dimension as a binary label, we computed precision, recall, and F1 score per dimension, along with macro-averaged scores across all dimensions, following standard multi-label evaluation practices~\cite{madjarov2012extensive, tsoumakas2008multi}.

%

\subsubsection{Trustor and Reason of Trust} 
LLMs were prompted to assign one trustor and one reason per post, separately, with results evaluated against our test set using precision, recall, and F1 (see Appendix~\ref{sec:app-prompts} for prompts).

%% file: 05-results.tex
\section{Results} \label{sec:result} 
\begin{figure}[t]
    \centering
\includegraphics[width=0.8\columnwidth]{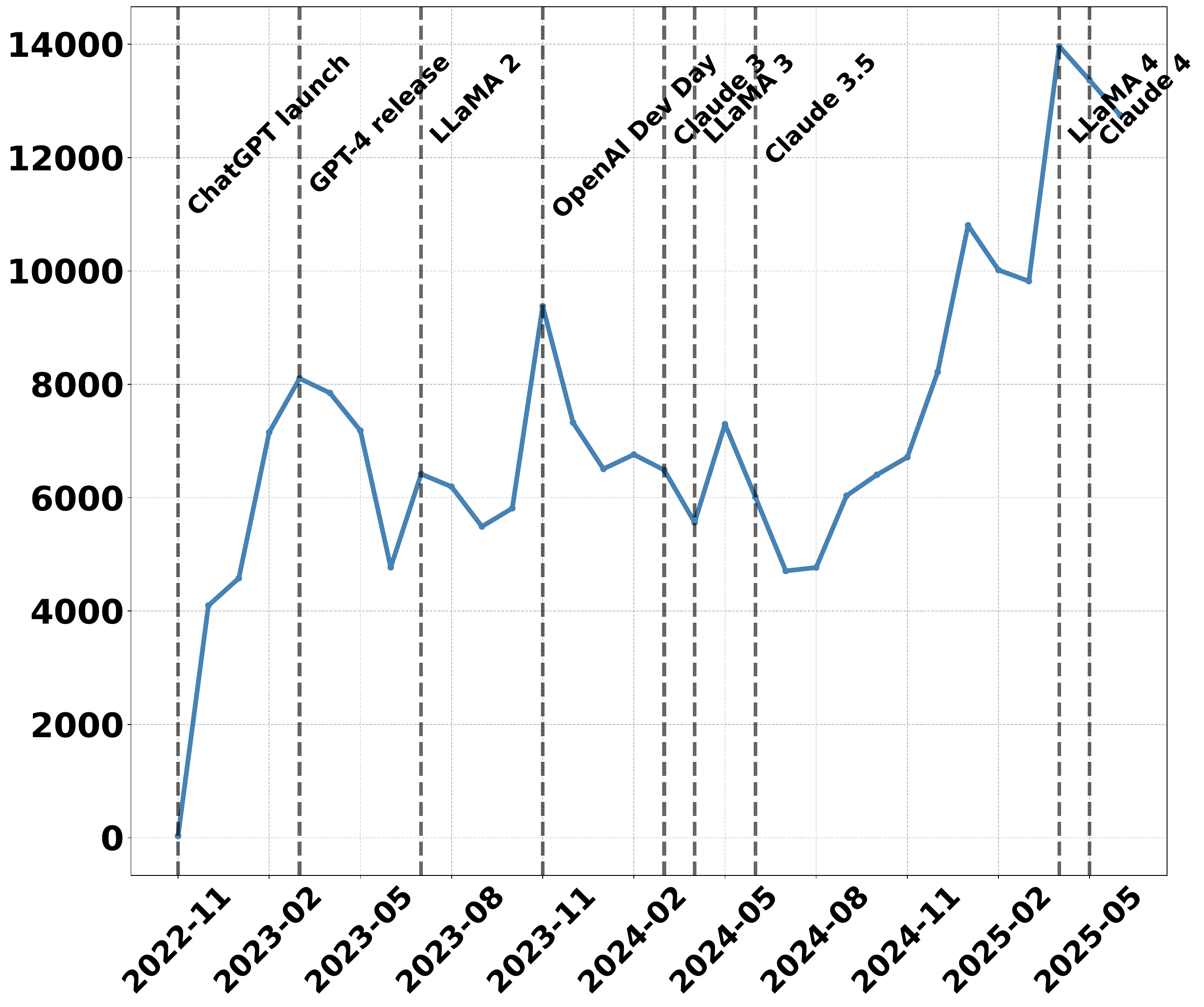}
    \caption{Temporal changes in the number of posts}
    \label{fig:number of posts}
    \vspace{-6mm}
\end{figure}

\noindent\textbf{Volume of Discourse on Reddit.}
Figure~\ref{fig:number of posts} shows how the volume of Reddit discourse about GenAI evolved between November 30, 2022, and June 30, 2025. The $x$-axis represents time (monthly bins), and the $y$-axis represents the number of posts containing GenAI keywords (after preprocessing). Overall, we observe a gradual but sustained increase in activity, punctuated by sharp peaks that align with major model announcements.  
The initial surge in late 2022 coincides with the public release of ChatGPT, which brought LLMs into mainstream public attention. Subsequent spikes correspond to milestone releases such as GPT-4 (March 2023), LLaMA2 (July 2023), and OpenAI's Dev Day (November 2023). While discussion levels fluctuated throughout 2023 and early 2024 despite new releases such as Claude3 (March 2024) and LLaMA3 (April 2024), the sharp and sustained growth at the end of 2024 into 2025 culminated in the highest levels of discourse during the releases of models such as LLaMA4 and Claude4 (April-May 2025). Although volume fluctuations may reflect multiple factors, the alignment of peaks with well-publicized model releases suggests that breakthrough announcements play a key role in driving community engagement.  

\begin{table}[t]
\centering
\resizebox{0.8\linewidth}{!}{%
\begin{tabular}{clccc}
\toprule
\multicolumn{1}{l}{Type} &
  Model &
  Precision &
  Recall &
  F1 \\
\midrule 
&
  BERT 
  &0.77 &0.75 &0.68 \\
 &
  RoBERTa 
  &0.76 & 0.79 & 0.77 \\
 &
  DistillBERT 
  &0.74 & 0.78 &0.75 \\
 &
  ModernBERT 
  & 0.74 & 0.76 & 0.75 \\
\multirow{-6}{*}{\rotatebox{90}{Transformers}} &
  SocBERT 
  &0.73 &0.74 & 0.69 \\
  \midrule
 &
  GPT-5.2 
  & 0.8 &	0.78 & 0.79 \\
 &
  GPT-5.1 
  & 0.81 & 0.8 &	\textbf{0.8} \\
 &
  GPT-5 
  & 0.8 &	0.74 & 0.76 \\
 &
  GPT-5-mini 
  & 0.8 &	0.79 & 0.79 \\
 &
 GPT-5-nano 
 & 0.79 & 0.77 &	0.78 \\
 &
   GPT-4.1 
  & 0.79 & 0.74 &	0.74 \\
 &
   GPT-4.1-mini 
  & 0.79 & 0.77 & 0.77 \\
 &
   GPT-4.1-nano 
  & 0.77 & 0.66 & 0.66 \\
 &
 GPT-4o 
  & 0.78 & 0.74& 0.75 \\
 &
  GPT-4o-mini 
  & 0.8 &	0.71 & 0.73 \\
 &
 GPT-oss-20b 
 & 0.79 & 0.75 & 0.76 \\
 &
  Mixtral-8x7B  
  & 0.8 & 0.56 & 0.62 \\

\multirow{-14}{*}{\rotatebox{90}{Zero-Shot}} &
  Llama-4-Maverick 
  & 0.76 & 0.76 & 0.75 \\
  \midrule
 &
  GPT-5.2 
  & 0.81 & 0.76 & 0.78 \\
 &
  GPT-5.1 
  & 0.79 & 0.77 &	0.77 \\
 &
  GPT-5 
  & 0.81 & 0.75 & 0.77 \\
 &
  GPT-5-mini 
  & 0.81 & 0.79 & \textbf{0.8} \\
 &
 GPT-5-nano 
 & 0.79 & 0.77 &	0.78 \\
 &
   GPT-4.1 
  & 0.81 & 0.75 &	0.77 \\
 &
   GPT-4.1-mini 
  & 0.8 & 0.78 & 0.79 \\
 &
   GPT-4.1-nano 
  & 0.76 & 0.61 & 0.62 \\
 &
 GPT-4o 
  & 0.8 & 0.79 & 0.79 \\
 &
  GPT-4o-mini 
  & 0.81 & 0.75 & 0.77 \\
 &
 GPT-oss-20b 
 & 0.81 & 0.76 & 0.78 \\
 &
  Mixtral-8x7B  
  & 0.79 & 0.37 & 0.47 \\

\multirow{-14}{*}{\rotatebox{90}{Few-Shot}} &
  Llama-4-Maverick 
  & 0.79 & 0.74 & 0.75 \\
\bottomrule
\end{tabular}}%
\caption{Model performance in trust classification}
\label{tab:Classification task 1}
\vspace{-5mm}
\end{table}

\subsection{Trust and Distrust}

\noindent\textbf{Classification Results.}
Table~\ref{tab:Classification task 1} reports performance across transformers, zero-shot, and few-shot LLMs for classifying \textit{Trust}, \textit{Distrust}, \textit{Both}, and \textit{Neither}. Among transformers, RoBERTa performed best (F1 = 0.77), with ModernBERT and DistillBERT close behind (F1 = 0.75), reflecting the advantages of more recent pretraining. Zero-shot LLMs generally outperformed transformers: GPT-5.1, GPT-5-mini, and GPT-5.2 reached F1 = 0.8, 0.79, and 0.79, respectively.  Open-source LLMs such as Llama-4-Maverick (F1 = 0.75) lagged. Few-shot LLMs showed a consistent performance, with GPT-5-mini (F1= 0.8), GPT-4.1-mini (F1= 0.79), and GPT-4o (F1= 0.79) leading.
The best overall performance was achieved by GPT-5.1 in the zero-shot setting, and GPT-5-mini in the few-shot setting (F1 = 0.8). Our results suggest that adding shots does not necessarily improve performance in this classification task.

Table~\ref{tab:performance-label} breaks down GPT-5.1's zero-shot performance for each category. Performance was strongest on \textit{Distrust} (F1 = 0.89) and \textit{Trust} (F1 = 0.84), indicating that the model reliably captured clear positive or negative context. Performance droped for \textit{Neither} (F1 = 0.60), showing difficulty in distinguishing neutral commentary from implicit evaluations. The lowest scores occurred for \textit{Both} (F1 = 0.24), highlighting the challenge in detecting ambivalent or ironic posts (see Appendix \ref{sec:app:error-analysis} for error analysis).
Based on these results and the overall performance of GPT-5.1-2025-11-13 (zero-shot), we used this model to generate labels for the full dataset with OpenAI's batch API.

\begin{table}[t]
\centering
\resizebox{0.7\columnwidth}{!}{%
\begin{tabular}{lccc}
\toprule
\textbf{Label} & \textbf{Precision} & \textbf{Recall} & \textbf{F1} \\
\midrule
Distrust      & 0.86           & 0.91        & 0.89\\
Trust          & 0.88           & 0.80       & 0.84\\   
Neither        & 0.55          & 0.65       & 0.60\\
Both           & 0.50          & 0.15        & 0.24\\
\bottomrule
\end{tabular}}
\caption{Performance of the best model (GPT-5.1 \textit{zero-shot}) in trust classification on each label
}
\label{tab:performance-label}
\end{table}

\begin{table}[t]
\centering
\resizebox{0.7\columnwidth}{!}{%
\begin{tabular}{lrrr}
\toprule
\textbf{Label} & \textbf{Annotated} & \textbf{All Data} & \textbf{\%} \\
\midrule
Neither & 369 & 95,513 & 41  \\
Trust & 812 & 71,431 & 31 \\ 
Distrust & 979 & 60,668 & 26 \\ 
Both & 67 & 2962 & 1 \\ 
\bottomrule
\end{tabular}}
\caption{Distribution of labels in annotated posts.}
\label{tab:annotation}
\vspace{-4mm}
\end{table}

\noindent\textbf{Label Distribution.}
Table~\ref{tab:annotation} presents the distribution of labels in the annotated subset and the full dataset. The largest category is \textit{Neither} (41\%), indicating that many GenAI-related posts are descriptive or informational (e.g., sharing news or announcements). \textit{Trust} (31\%) and \textit{Distrust} (26\%) are nearly balanced, reflecting the polarized nature of public opinion. By contrast, \textit{Both} (1\%) is marginal, suggesting that ambivalent expressions are comparatively rare and harder to detect.

\begin{figure}[ht]
    \centering

        \includegraphics[width=\columnwidth]{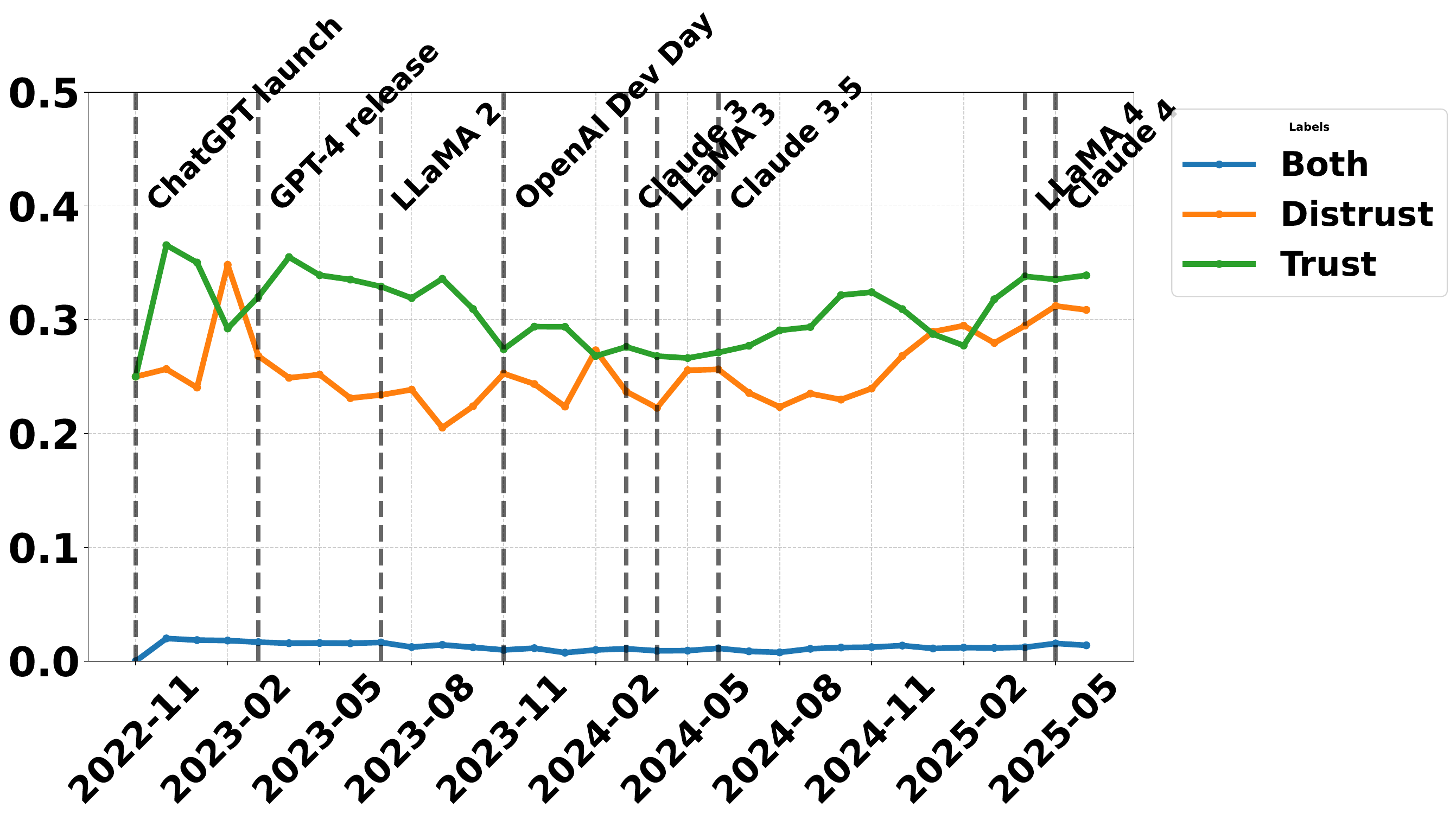}
    \caption{Comparison between Trust vs Distrust.}
    \label{fig:trust_distrust_combined}
\end{figure}

\noindent\textbf{Temporal Changes in Trust \& Distrust.}
Figure~\ref{fig:trust_distrust_combined} shows the monthly changes in the number of posts labeled as \textit{Trust}, \textit{Distrust}, and \textit{Both}, normalized by the total number of posts in that month. The figure suggests that `Trust' consistently, but modestly, outweighs `Distrust' across the timeline of our data. While \textit{Trust} accounts for roughly 30–35\% of posts across most months, \textit{Distrust} appears in roughly 20-30 \% of posts.
\textit{Both}, however, accounts for a very insignificant share of posts (We did not add \textit{Neither} to the figure, but it presents the largest number of posts in our data). We observe peaks around major model releases, while \textit{Trust} rises in mid-2023 following GPT-4 and LLaMA~2, \textit{Distrust} spikes around OpenAI's Dev Day (late 2023). Despite these fluctuations, \textit{Trust} maintains its modest lead over \textit{Distrust}. However, neither category dominates the discourse, underscoring a persistent tension in public attitudes toward GenAI~\cite{marantz2024doomsayers, edinburgh2023doomersboomers, acm2023doomerboomer}. The flat trend for \textit{Both} indicates that ambivalence is rare compared to strong positive or negative stances.


\begin{table*}[t]
\centering
\resizebox{0.9\textwidth}{!}{%
\begin{tabular}{cclccccccccccccccc}
\toprule
Strategy &
\multicolumn{1}{l}{Type} &
  Model &
  \multicolumn{13}{c}{Dimensions} &
  \multicolumn{2}{c}{Average} \\
  \cline{4-16}
\multicolumn{1}{l}{} &\multicolumn{1}{l}{} &&
  Ben &
  Com &
  Dec &
  Dis &
  Fam &
  Inc &
  Int &
  Mal &
  Opa &
  Rel &
  Tra &
  Unf &
  Unr &
  \multicolumn{1}{c}{\begin{tabular}[c]{@{}c@{}}F1\end{tabular}} \\
  \hline
\multirow{9}{*}{Majority} &
\multirow{4}{*}{Zero-Shot} &

Llama-4 &
  0.36 & \textbf{0.89} & 0.45 & 0.41 & 0.25 & 0.75 & 0.22 & 0.51 & 0.27 & 0.43 & 0.19 & 0.14 & 0.74 & 0.43 \\
&
&  gpt-5-2025-08-07 &
  0.32 & 0.88 & 0.43 & 0.42 & 0.32 & 0.74 & \textbf{0.29} & 0.52 & 0.28 & 0.42 & 0.20 & 0.11 & 0.67 & 0.43 \\
&
&  gpt-5-mini-2025-08-07 &
  0.36 & \textbf{0.89} & 0.42 & 0.45 & 0.31 & \textbf{0.76} & 0.25 & 0.55 & 0.34 & 0.68 & 0.20 & 0.11 & 0.79 & 0.47 \\
&
&  gpt-4o-2024-11-20 &
  0.35 & \textbf{0.89} & 0.45 & 0.39 & 0.30 & 0.73 & 0.20 & 0.43 & \textbf{0.39} & 0.46 & 0.14 & 0.15 & 0.77 & 0.43 \\\cline{2-18}
&\multirow{4}{*}{Few-Shot} &  gpt-5.1-2025-11-13 &
  0.38 & 0.88 & 0.47 & 0.36 & 0.32 & 0.71 & 0.19 & 0.41 & 0.32 & \textbf{0.69} & \textbf{0.24} & 0.17 & 0.77 & 0.45 \\
&
&  gpt-5-mini-2025-08-07 &
  0.36 & \textbf{0.89} & 0.45 & 0.42 & \textbf{0.34} & 0.73 & 0.25 & 0.51 & 0.34 & 0.57 & 0.19 & 0.15 & \textbf{0.80} & 0.46 \\
&
&  gpt-4o-mini-2024-07-18 &
  \textbf{0.45} & 0.88 & 0.44 & \textbf{0.51} & 0.30 & 0.73 & 0.20 & 0.44 & 0.35 & 0.57 & 0.15 & 0.13 & 0.78 & 0.46 \\
&
&  gpt-4.1-2025-04-14 &
  0.43 & 0.88 & \textbf{0.49} & 0.37 & \textbf{0.34} & 0.75 & 0.23 & 0.44 & 0.38 & 0.67 & 0.20 & 0.12 & \textbf{0.80} & 0.47 \\

\hline
\multirow{10}{*}{Any} &
\multirow{5}{*}{Zero-Shot} &
Mixtral-8x7B &
  0.47 & 0.95 & 0.35 & 0.45 & 0.70 & 0.48 & 0.12 & 0.33 & 0.53 & 0.56 & 0.25 & 0.15 & \textbf{0.93} & 0.48 \\
&
& Llama-4 &
  0.32 & \textbf{0.97} & 0.56 & 0.42 & 0.30 & 0.86 & 0.23 & 0.42 & 0.20 & 0.41 & 0.11 & 0.09 & 0.75 & 0.43 \\
&
&  gpt-5-mini-2025-08-07 &
  0.41 & \textbf{0.97} & 0.53 & \textbf{0.52} & 0.73 & \textbf{0.87} & 0.20 & 0.36 & 0.52 & 0.68 & 0.23 & 0.23 & 0.84 & 0.55 \\
&
&  gpt-4o-2024-11-20 &
  0.43 & \textbf{0.97} & \textbf{0.59} & 0.45 & 0.75 & 0.77 & 0.10 & 0.53 & 0.44 & 0.44 & 0.13 & 0.20 & 0.82 & 0.51 \\
&
&  gpt-4.1-2025-04-14 &
  0.29 & 0.96 & 0.52 & 0.36 & \textbf{0.77} & 0.81 & 0.10 & 0.49 & 0.48 & 0.58 & 0.09 & 0.30 & 0.85 & 0.51 \\\cline{2-18}
&
\multirow{5}{*}{Few-Shot} &
 Mixtral-8x7B &
  \textbf{0.57} & 0.95 & 0.22 & 0.51 & 0.68 & 0.71 & \textbf{0.26} & 0.46 & 0.55 & 0.66 & 0.35 & 0.23 & 0.90 & 0.54 \\
&
&  gpt-5-2025-08-07 &
  0.26 & \textbf{0.97} & 0.40 & 0.41 & 0.58 & 0.78 & 0.16 & 0.42 & 0.45 & 0.47 & 0.21 & 0.18 & 0.74 & 0.46 \\
&
&  gpt-5-mini-2025-08-07 &
  0.40 & \textbf{0.97} & 0.49 & 0.44 & 0.72 & 0.72 & 0.17 & 0.44 & \textbf{0.57} & 0.56 & 0.26 & 0.18 & 0.83 & 0.52 \\
&
&  gpt-4.1-2025-04-14 &
  0.39 & 0.96 & 0.52 & 0.36 & 0.61 & 0.74 & 0.14 & 0.53 & 0.50 & \textbf{0.74} & 0.12 & 0.18 & 0.83 & 0.51 \\
&
&  gpt-4.1-mini-2025-04-14 &
  0.54 & 0.92 & 0.47 & 0.24 & 0.62 & 0.61 & 0.10 & \textbf{0.60} & 0.38 & 0.63 & 0.20 & 0.16 & 0.76 & 0.48 \\
\bottomrule
\end{tabular}}
\caption{Selected classification result of dimensions by zero-shot and few-shot models, and any and majority strategies. Dimensions: Ben: Benevolence, Com: Competence, Dec: Deception, Dis: Dishonesty, Fam: Familiarity, Inc: Incompetence, Int: Integrity, Mal: Malevolence, Opa: Opaqueness, Rel: Reliability, Tra: Transparency, Unf: Unfamiliarity, Unr: Unreliability. Full result is shared in Tables \ref{tab:majority_task_style} and \ref{tab:any_task_style}.
}
\label{tab:any_majority_task2}
\end{table*}

\subsection{Dimensions of Trust and Distrust}

\noindent\textbf{Classification Results.}
%
Table \ref{tab:any_majority_task2} shows the selected results of the dimension classification (full result in Appendix, Table~\ref{tab:majority_task_style}). Under the \textit{majority} condition, using zero-shot promoting, models achieved their highest scores on functionality-related and explicit dimensions, particularly Competence, Incompetence, and Unreliability, where GPT-5 variants and Llama-4 obtained the highest scores overall. In contrast, interpersonal and normative dimensions such as Benevolence, Integrity, and especially Transparency remained comparatively weak, with low peak scores across all models. In the few-shot setting, improvements were most evident for relational dimensions including Reliability, Dishonesty, and Familiarity, where GPT-4.1 and GPT-5 showed higher and more stable performance. However, even with few-shot prompting, dimensions capturing ethical aspects e.g., notably Integrity and Transparency, continued to exhibit limited classification performance relative to other categories.
Under the \textit{any} criterion (full result in Appendix, Table~\ref{tab:any_task_style}), overall scores increased in both prompting settings. In the zero-shot, models again performed best on Competence and Incompetence, with several models reaching almost perfect performance, while Unreliability also achieved consistently high scores across models. Reliability improved substantially under few-shot prompting, where GPT-4.1–series models achieved the highest scores. In contrast, Deception and Dishonesty were slightly better classified with zero-shot, suggesting that these dimensions are less dependent on additional contextual examples. Dimensions such as Integrity and Transparency remained comparatively difficult to classify across both techniques, indicating persistent difficulty in detecting more abstract or implicitly expressed dimensions.

Based on the results and comparing recall, percision and F1 score of the models, we observe that the \textit{majority} approach penalizes over-prediction of dimensions, rewarding consensus. On the other hand, the \textit{any} approach rewards identifying dimensions that are more challenging to detect, encouraging the diversity in how dimensions of trust or trust can be implied. 
Given the variability in performance across dimensions and the absence of a single model that consistently outperformed the others, we selected the best-performing model for each dimension based on the \textit{any} evaluation results. We then applied this dimension-specific model selection strategy to annotate the full dataset.


\begin{table}[t]
\centering
\resizebox{0.8\columnwidth}{!}{%
\begin{tabular}{clcc}
\toprule
&\textbf{Dimension}     & \textbf{Annot (Majority/Any)} & \textbf{All Data} 
\\
\midrule
&Competence    & 747/ 1095     &  70,851 
\\
&Reliability   & 611/ 1068     &  48,011 
\\
\multirow{-1}{*}{\rotatebox{90}{Trust}}&Familiarity   & 176/ 793      &  57,496  
\\
&Transparency  & 82/ 649      &  35,514 
\\
&Benevolence   & 61/ 416      &  21,887 
\\
&Integrity     & 58/ 580      &   8,480  
\\
\cmidrule(lr){1-4}
&Unreliability & 689/ 1179     &  56,257 
\\
&Incompetence  & 563/ 1081     &  42,312 
\\
&Opaqueness    & 114/ 700      &  24,407 
\\
&Deception     & 159/ 676      &  23,478 
\\
&Dishonesty    & 156/ 709      &  17,631   
\\
\multirow{-4}{*}{\rotatebox{90}{Distrust}}&Unfamiliarity & 23/ 454      &   24,469 
\\
&Malevolence   & 60/ 332      &  10,303 
\\
\bottomrule      
\end{tabular}}
\caption{Dimension frequencies in the annotated sample and entire data, each post can have one or more dimensions.}
\label{tab:Dimensions}
\vspace{-4mm}
\end{table}

\noindent\textbf{Label Distribution.}
Table~\ref{tab:Dimensions} reports the frequency of dimensions in the annotated set (for two ground-truth methods) and the entire dataset, predicted by LLMs. Within \textit{Trust}, \textit{Competence}, \textit{Reliability}, and \textit{Familiarity} are most common, reflecting how users primarily evaluate GenAI in terms of technical performance and personal experience. By contrast, ethical or normative dimensions such as \textit{Integrity}, \textit{Transparency}, and \textit{Benevolence} appear less frequently. Within \textit{Distrust}, \textit{Unreliability}, and \textit{Incompetence} dominate, echoing concerns about functionality and accuracy. Other dimensions, such as \textit{Deception}, \textit{Dishonesty}, and \textit{Opaqueness}, occur moderately often, while explicitly value-focused categories like \textit{Malevolence} remain rare.
Overall, the functionality-related and performance-oriented dimensions dominate both \textit{Trust} and \textit{Distrust} discourse, while ethical and normative aspects are less present.


\begin{figure*}[t]
    \centering

    \begin{subfigure}[t]{0.45\textwidth}
        \centering
        \includegraphics[width=\columnwidth]{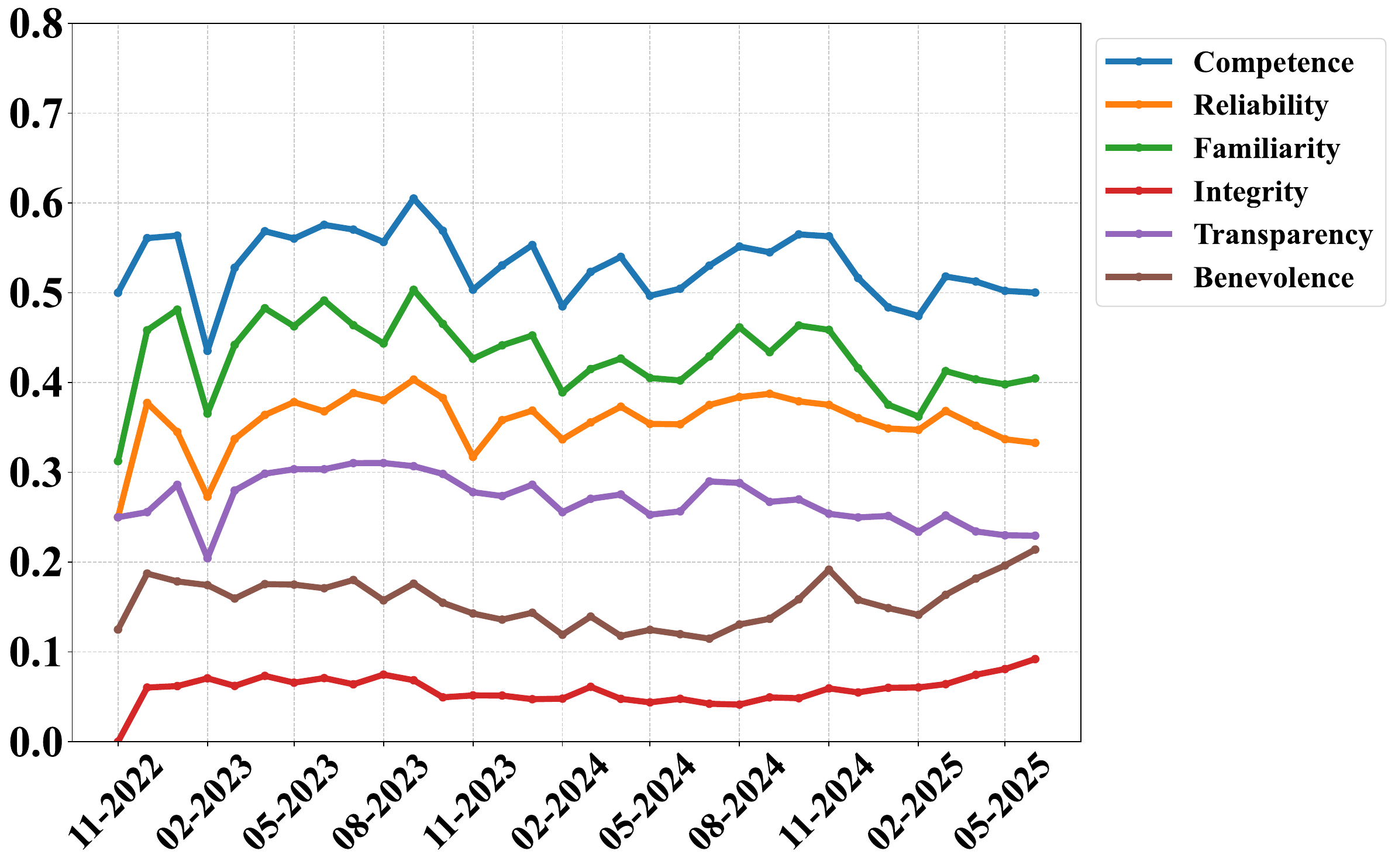}
        \caption{Dimensions of Trust}
        \label{fig:Dimensions of Trust}
    \end{subfigure}
    \begin{subfigure}[t]{0.45\textwidth}
        \centering
        \includegraphics[width=\columnwidth]{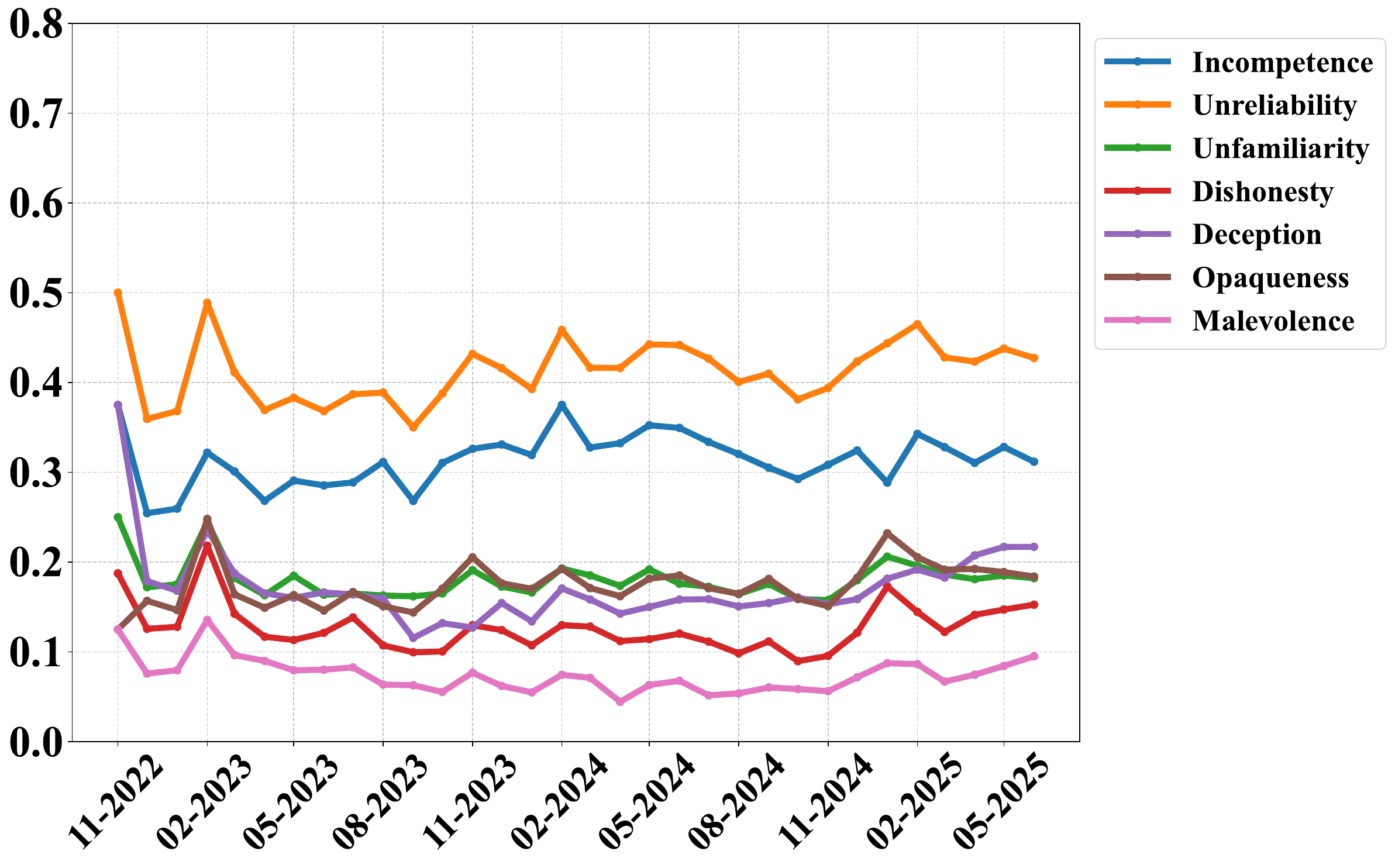}
        \caption{Dimensions of Distrust}
        \label{fig:Dimensions of Distrust}
    \end{subfigure}

    \caption{Normalized Distribution of Dimensions}
    \label{fig:Dimensions}
\end{figure*}

\noindent\textbf{Temporal Changes of Dimensions.}
Figure~\ref{fig:Dimensions} shows how dimensions evolve in our data.
Among dimensions of \textit{Trust}, \textit{Competence} is the most dominant, maintaining the highest prevalence across the entire timeline. This suggests that perceptions of whether GenAI can effectively perform tasks are the central lens through which trust is framed. \textit{Familiarity} also remains consistently high, indicating that repeated exposure and growing experience with these systems reinforce \textit{Trust} over time. \textit{Reliability} appears somewhat lower but still steady, pointing to its importance as a complementary indicator of technical soundness. \textit{Integrity}, \textit{Transparency}, and \textit{Benevolence} remain marginal throughout. 
Their relative flatness across the timeline shows that while ethical values are present in discourse, they play a secondary role compared to functionality and user experience in social media discourse. The \textit{Distrust} dimensions show greater volatility and relatively lower presence in the data. \textit{Unreliability} and \textit{Incompetence} dominate this category, reflecting sustained concerns with system accuracy and task performance. 
\textit{Opaqueness} and \textit{Deception} also show modest rises over time, which can point to ongoing worries about transparency and hallucination. In contrast, \textit{Malevolence} and \textit{Dishonesty} remain minimal throughout, rarely gaining traction even in periods of heightened public debate. 

\noindent\textbf{Dimensions Co-occurence.}
To understand the relationships between dimensions, we calculated the \emph{Jaccard index} for co-occurrence of different pairs of dimensions (Table~\ref{tab:Most Co-occurrent Pairs of Dimensions} in Appendix).
In the \textit{Trust} posts, the strongest co-occurrence is between \textit{Competence} and \textit{Familiarity} (J = 0.76), followed by \textit{Competence} with \textit{Reliability} (0.66) and \textit{Familiarity} with \textit{Reliability} (0.52).  Pairs such as \textit{Competence}-\textit{Transparency} and \textit{Transparency}-\textit{Familiarity} show moderate overlap, suggesting that technical performance and clarity of system behavior are often discussed together.
For \textit{Distrust}, the dominant pair is \textit{Incompetence}-\textit{Unreliability} (J = 0.67), indicating that functionality failures are central to negative perceptions. Other notable co-occurrences were \textit{Opaqueness}-\textit{Unreliability}, and \textit{Unreliability}-\textit{Deception}, reflecting concerns about opacity and misinformation. Overall, Jaccard was higher for pairs of trust dimensions compared to distrust dimensions. 



\begin{table}[t]
\centering

\label{tab:combined_metrics}
\resizebox{0.8\columnwidth}{!}{%
\begin{tabular}{lccc}
\toprule
\textbf{Model} & \textbf{Precision} & \textbf{Recall} & \textbf{F1-Score} \\
\midrule
\multicolumn{4}{c}{\textit{Reasons of Trust}} \\
\midrule
GPT-5.2 & 0.90 & 0.80 & 0.85 \\
GPT-5.1 & 0.86 & 0.75 & 0.79 \\
GPT-5 & 0.90 & 0.82 & 0.85 \\
GPT-5-mini & 0.91 & 0.85 & \textbf{0.87} \\
GPT-5-nano & 0.91 & 0.77 & 0.82 \\
GPT-4o & 0.91 & 0.78 & 0.83 \\
GPT-4o-mini & 0.87 & 0.71 & 0.78 \\
GPT-4.1 & 0.89 & 0.85 & 0.86 \\
GPT-4.1-mini & 0.88 & 0.79 & 0.83 \\
GPT-4.1-nano & 0.85 & 0.15 & 0.22 \\
\midrule
\multicolumn{4}{c}{\textit{Trustors}} \\
\midrule
GPT-5.2 & 0.80 & 0.79 & 0.78 \\
GPT-5.1 & 0.81 & 0.82 & \textbf{0.80} \\
GPT-5 & 0.80 & 0.79 & 0.78 \\
GPT-5-mini & 0.79 & 0.81 & 0.79 \\
GPT-5-nano & 0.78 & 0.78 & 0.78 \\
GPT-4o & 0.80 & 0.79 & 0.79 \\
GPT-4o-mini & 0.77 & 0.79 & 0.76 \\
GPT-4.1 & 0.80 & 0.81 & 0.80 \\
GPT-4.1-mini & 0.78 & 0.81 & 0.78 \\
GPT-4.1-nano & 0.69 & 0.67 & 0.67 \\
\bottomrule
\end{tabular}%
}
\caption{Classification for Reason and Trustor}
\label{tab:Reason Trustor Classification}
\vspace{-0.5cm}
\end{table}

\noindent\textbf{Classification Performance vs. Human Agreement}
We noticed that the classification performance was lower than expected for several dimensions, including \textit{Integrity, Transparency}, and \textit{Unfamiliarity}, even with their best respective models. To further examine this, we analyzed how human annotators employed different dimensions during coding. We computed an aggregate ratio for each dimension reflecting both their frequency of use and the level of annotator agreement. Specifically, the ratio corresponds to the proportion of annotators assigning a given dimension to a post. For instance, if three out of five annotators label a post with Competence, the ratio for that dimension is 3/5, whereas dimensions not selected by any annotator receive a value of zero. These ratios were then aggregated across the annotated dataset for each dimension (Table \ref{tab:dimension_means_small}).
To assess classification performance across dimensions, we examined the relationship between aggregate annotation ratios and classification scores under the \textit{majority} and \textit{any} evaluation criteria.
Our results (Figure \ref{fig:classification_agreement} in Appendix) suggest a strong Pearson correlation between aggregate rates and classification performance (r= 0.83 in both approaches and statistically significant). While functionality-related dimensions such as \textit{Competence, Reliability} and their semantic opposites demonstrate both a high agreement rate among human annotators and a high classification performance, ethical dimensions like \textit{Integrity} and \textit{Transparency} are suffering from both a low aggregated presence in human-annotators labels and a low classification performance.



\subsection{Reason \& Trustor}
\noindent\textbf{Classification Results.}
Table~\ref{tab:Reason Trustor Classification} reports model performance on identifying the implied \textit{Reason} for \textit{Trust} or \textit{Distrust}, as well as the \textit{Trustor} in each post.
Performance was relatively strong across models for both tasks. For \textit{Reason}, GPT-5-mini performed best (F1 = 0.87), surpassing other models on both precision and recall.
For \textit{Trustor}, F1 scores were very close. GPT-5.1 and GPT-4.1 achieved the highest F1 (0.8), though GPT-5.1 showed slightly higher recall and precision. Given these results, we adopted GPT-5-mini for \textit{Reason} and GPT-5.1 for \textit{Trustor} to label the rest of our dataset with batch processing.

\begin{figure}[t]
\centering
\includegraphics[width=0.8\columnwidth]{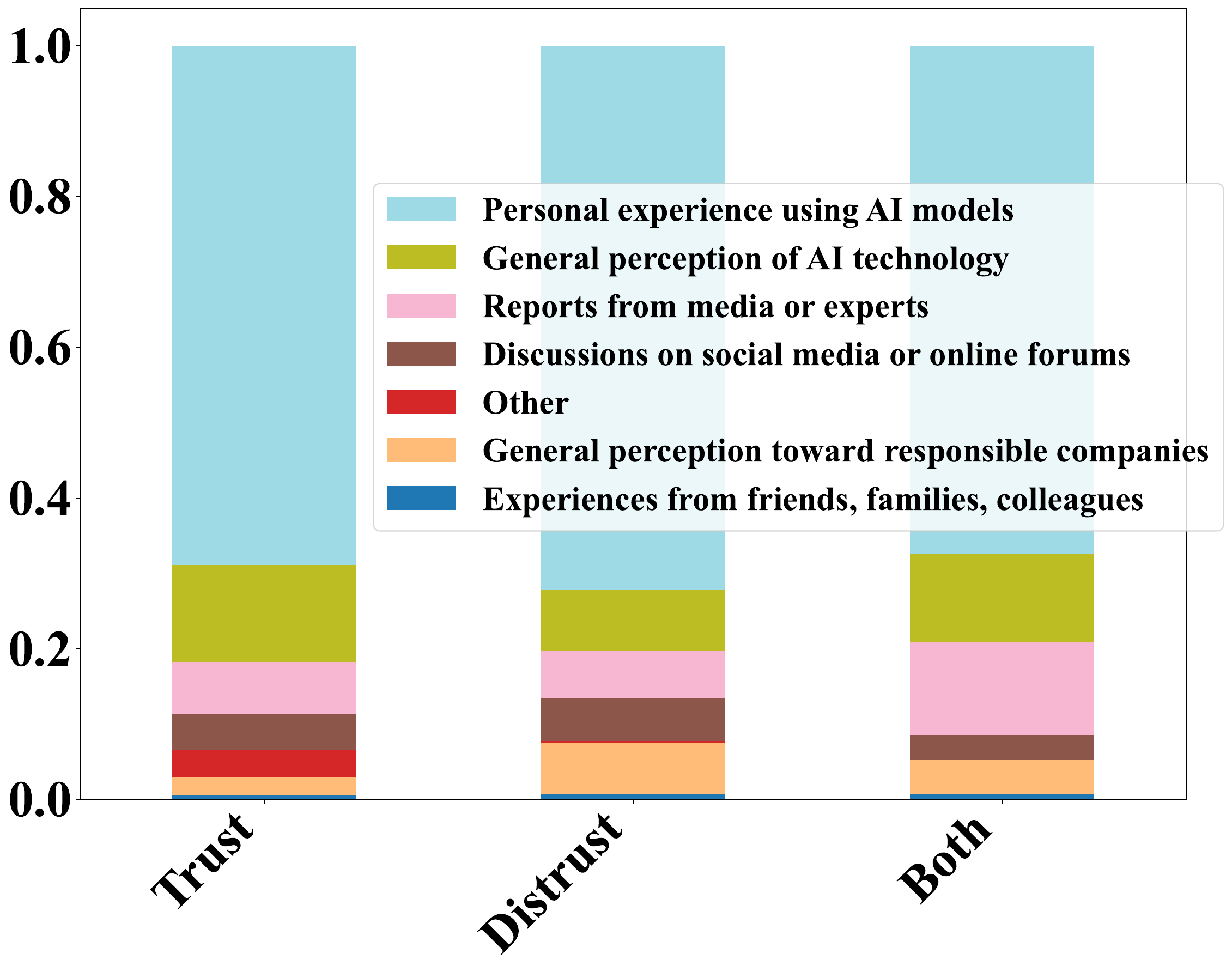}
\caption{Distribution of Reasons}
\label{fig:reason_and_labels}
\vspace{-5mm}
\end{figure}

\noindent\textbf{Reasons of Trust.}
Figure~\ref{fig:reason_and_labels} shows the distribution of \textit{reasons} cited for posts labeled as \textit{Trust}, \textit{Distrust}, and \textit{Both}. Across all three categories, personal experience using AI accounts for the majority of posts, indicating that direct interaction with GenAI was the primary driver of both \textit{Trust} and \textit{Distrust}.
Among secondary reasons, general perception of AI appeared more frequently in \textit{Trust} posts, while references to responsible companies were more common in \textit{Distrust}. Media or experts' reports were cited as the secondary reason for posts tagged as \textit{Both}. 
Mentions of social media discussions, peer influence, or family experiences were minimal, suggesting these external factors play only a supplementary role. Overall, these results reinforce that trust and distrust in GenAI are predominantly framed through firsthand use, with external narratives adding nuance but less weight \cite{openai2025howpeopleareusingchatGPT}.

\begin{figure}[t]
\centering
\includegraphics[width=\columnwidth]{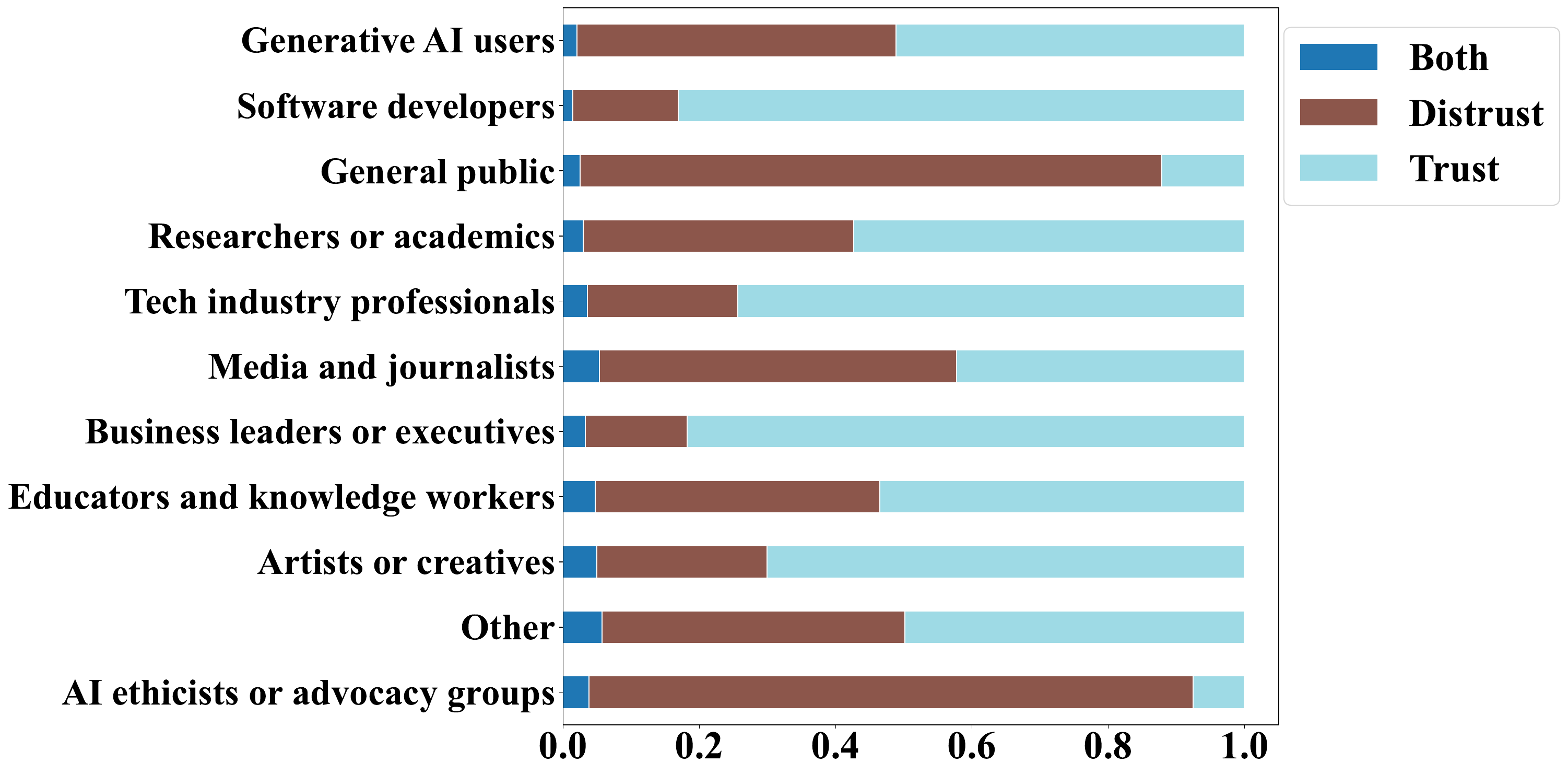}
\caption{Distribution of different Trustor groups}
\label{fig:comparison_trustor}
\vspace{-5mm}
\end{figure}

\noindent\textbf{Trustors.} Figure~\ref{fig:comparison_trustor} shows the breakdown of labels within each \textit{Trustor} group. Among business leaders, academics, software developers, and tech professionals, \textit{Trust} outweighs \textit{Distrust}. In contrast, \textit{Distrust} is more prevalent among AI ethicists, the general public, media \& journalists (although smaller). Generative AI users (the largest group), and educators \& knowledge workers were distinctive in showing relatively balanced levels of \textit{Trust} and \textit{Distrust}, suggesting a more cautious or divided perspective.
These patterns show that while direct user experience drives the overall framing of (dis)trust, different stakeholders emphasize contrasting evaluations, highlighting the sociotechnical complexity of GenAI discourse.

%% file: 06-discussion.tex
\section{Discussion} \label{sec:discussion}

\noindent\textbf{Trust and Distrust Are Not Static. }
Our analysis shows that \textit{Trust} and \textit{Distrust} coexist in GenAI discourse, each accounting for a substantial share of posts, with local surges triggered by model releases. This dynamic mirrors survey findings that public attitudes toward AI remain mixed, balancing optimism with concern~\cite{pew2025_us_public_ai_experts}. It echoes work showing that breakthrough announcements coincide with intensified debate and media coverage~\cite{ng2024powerful, leiter2024chatgpt}. Rather than converging toward stable positions, online discourse repeatedly oscillates between optimism and skepticism, showing the evolving nature of debates about the technology's value and risks. This is consistent with process-oriented models of human–AI trust in the literature \cite{mcgrath2025collaborative}.
Dimensions analysis reveals that discourse is anchored in functionality: \textit{Competence}, \textit{Reliability}, and \textit{Familiarity} dominate, while ethical dimensions such as \textit{Transparency}, \textit{Integrity}, and \textit{Benevolence} are less salient. This imbalance suggests that users prioritize whether GenAI ``works'' over whether it is ``good,'' echoing prior findings that functionality and reliability form the baseline for technology adoption~\cite{jiang2025understanding, gefen2024evolving}. Ethical concerns, though visible, remain secondary in everyday discussions, surfacing primarily in moments of controversy or policy debate~\cite{crawford2021atlas}. Dimension co-occurrence analysis shows that functional dimensions frequently appear together. Jaccard values are higher for trust pairs than for distrust pairs, suggesting users often combine multiple positive aspects of trustworthiness, while expressing distrust more narrowly.

\noindent\textbf{Dimensions: Classification vs. Salience.}
The relationship between classification performance and aggregate annotation ratios (which reflect both dimension prevalence and annotator agreement) indicates that model performance is closely tied to the clarity and consistency with which dimensions are expressed. Dimensions such as \textit{Competence} and \textit{Reliability}, which exhibit higher aggregate ratios, also achieve stronger classification performance. This suggests that models perform best when signals are both frequent and consistently interpreted by annotators.
In contrast, dimensions such as \textit{Integrity} and \textit{Transparency} appear less frequently and show weaker classification results. These dimensions are more abstract and value-driven, making them harder for both annotators and models to identify reliably. The observed performance gap, therefore, reflects not only computational limitations but also differences in how users articulate (dis)trust online: concrete performance-related concerns tend to be expressed explicitly, whereas ethical critiques often require broader contextual interpretation~\cite{binns2018s}.
Prior work shows that annotation disagreement in subjective tasks frequently reflects genuine interpretive variation rather than noise, and that model performance is constrained by the level of human consensus in the training data~\cite{aroyo2015truth, pavlick2019inherent}. Our findings suggest that lower performance on certain dimensions reflects variability in human interpretation as much as modeling limitations.

\noindent\textbf{Trust and Distrust Depend on Who.}
Trust in GenAI is not uniform but shaped by the identity and perspective of the trustor \cite{balayn2024empirical}.
Our analysis shows that business leaders, academics, software developers, media, and tech professionals express more \textit{Trust} than \textit{Distrust}, consistent with survey evidence that experts tend to be more optimistic about AI than the general public \cite{Pew2025PublicExpertAI}. By contrast, AI ethicists and the general public lean more toward \textit{Distrust}, reflecting critical perspectives grounded in ethical concerns and lived experience, echoing national survey findings of a widening AI optimism–skepticism divide \cite{Rutgers2025AI_divide}.
Educators and knowledge workers show almost equal levels of \textit{Trust} and \textit{Distrust}, suggesting a more cautious or divided perspective. This aligns with prior work showing ambivalence in professional settings where AI intersects with labor and pedagogy~\cite{buhmann2023deep}.
Results for \textit{Reasons} highlight personal experience as the dominant basis for \textit{Trust} or \textit{Distrust}, outweighting mentions of external factors such as media reports and AI companies. This is consistent with usage studies by OpenAI and Anthropic  \cite{openai2025_how_people_using_chatgpt, anthropic_claude_support_advice_companionship}, and previous research in HCI \cite{razi2024not}, which found most LLM interactions to be direct and experiential.

\noindent\textbf{Governance, Design, and Literacy.}
We show that \textit{Trust} and \textit{Distrust} coexist, and rather than converging toward a stable consensus, they continuously fluctuate. Governance frameworks and design practices must therefore treat both as independent forces, echoing the ``trust and distrust paradox'' where confidence and skepticism are expressed simultaneously~\cite{ou2009trust}. 
Our finding that functionality-oriented dimensions dominate over ethical ones raises questions about alignment with theoretical frameworks of \textit{Trust}. Previous work \cite{mayer1995integrative,glikson2020human,lee2004trust} incorporates both cognitive and affective factors that shape trust, emphasizing ability, integrity, and benevolence as fundamental foundations of \textit{Trust} that co-exist. Yet in public discourse, ability-based judgments dominate, while integrity and benevolence are far less salient. This suggests that classical definitions may not fully capture social media discourse on GenAI, how users articulate \textit{Trust} in GenAI, or that users implicitly collapse ethical and normative concerns into judgments of performance. This pattern likely reflects the experiential nature of social media discourse, where users evaluate technologies through direct interaction rather than institutional or ethical reasoning, overlooking normative, ethical concerns. Our results thus invite reconsideration of whether \textit{Trust} in AI, as expressed in social media narratives, aligns with or diverges from established theoretical frameworks.
From a digital literacy perspective, our findings highlight a gap between expert concerns and everyday user priorities: while policymakers emphasize fairness, bias, and accountability, most users focus on whether GenAI ``works'' in practice. Here, ``working'' often reflects perceived usefulness, whether outputs appear coherent, helpful, or time-saving, rather than evaluation of factual accuracy, bias, or completeness. A response that ``looks acceptable'' may still contain subtle errors or normative distortions that go unnoticed in routine use. Strengthening AI literacy therefore requires making normative risks visible in ways that connect to everyday use, ensuring governance frameworks address both practical utility and epistemic quality. These findings underscore the sociotechnical complexity of GenAI and the need for more comprehensive AI governance~\cite{long2020ai,jobin2019global}.

%% file: 08-ethics_limitation.tex
\section{Limitations} \label{sec:limitation} 
Our study has several limitations. First, in the classification of dimensions, performance varied across labels. Categories such as \textit{Competence} and \textit{Reliability} were classified with higher accuracy, but abstract dimensions like \textit{Transparency} and \textit{Integrity} performed poorly. We investigated the overall presence of dimensions in the sample set and found the link between the scarcity of these dimensions in the data and classification performance. We retained these dimensions but interpret them cautiously. Future work could improve reliability with expert input or hierarchical schemes that separate functional from normative dimensions. 
Second, our data is only in English and comes from Reddit, which, while large and temporally rich, skews younger and technologically engaged. Thus, our findings reflect specific online communities rather than the general population, and cross-platform studies are needed for broader validation.
While LLM-assisted classification is promising, models still struggle with ambivalence (\textit{Both}) and abstract normative categories, which are difficult to detect in informal posts.

\section{Ethical Considerations} \label{sec:ethics}

All data collection and annotation procedures were conducted under institutional IRB approval. Participants provided informed consent prior to beginning the study and were clearly informed of their rights, including the ability to withdraw at any time without penalty.
Annotators recruited through Prolific were compensated at an average rate of 10 USD/hour, which is above minimum wage and consistent with recommended fair-pay practices for crowdsourced research. We also included mechanisms to support participant well-being, such as a help button in the annotation interface and direct communication channels with the research team for resolving concerns.
All posts were collected from publicly accessible subreddits, and we excluded deleted or removed content to respect author intent. To minimize risks of re-identification, posts were annotated in de-identified form, and no usernames or direct links were shared. We recognize that research on trust and distrust in GenAI carries broader ethical stakes. Our findings describe how public perceptions evolve in online communities, but should not be misused as definitive judgments about the reliability or trustworthiness of GenAI systems themselves. Instead, this work contributes to transparent monitoring of public discourse, with the goal of informing responsible AI design, governance, and literacy initiatives.

%% file: 10-acknowledgement.tex
\section*{Acknowledgment} \label{sec:acknowledgment}

This research was supported by an award from Drexel University's Areas of Excellence \& Opportunity. We thank Farnaz Ghashami and Kshitij Kayastha for their contributions in the early stages of this project. We also thank Trevor Canfor-Dumas and Jesus Romero for their help with the literature review and annotation of sample data. 
Finally, we thank the Prolific participants whose time and engagement made this study possible.

%% file: 09-appendix.tex
\newpage
\section{Data Collection}\label{sec:app-data}

Table \ref{tab:Subreddit_Overview} shows the list of selected subreddits with counts of raw extracted posts and cleaned posts (before removing short posts). Here is the list of keywords related to generative AI:
\textit{Large Language Model*}, \textit{LLM}, \textit{LLMs}, \textit{LLM*}, \textit{GPT*}, \textit{ChatGPT*}, \textit{Google Bard}, \textit{LLaMA*}, \textit{Claude Anthropic}, \textit{AI apocalypse}, \textit{Claude}, \textit{OpenAI}, \textit{Gemini}, \textit{Language Models}, \textit{PaLM}, \textit{Pathways Language Model}, \textit{Fine-tuned Language Net}, \textit{FLAN}, \textit{LaMDA}, \textit{BLOOM}, \textit{Language Model for Dialogue Application}, \textit{BigScience Large Open-science Open-access Multilingual Language Model}, \textit{Mistral AI}, \textit{Mixtral}, \textit{Replika}, \textit{Deepseek}, \textit{Gen-AI}, \textit{GenAI}, \textit{generative ai}, \textit{Gen AI}, \textit{generative artificial intelligence}.

\begin{table}[h]
    \resizebox{0.9\columnwidth}{!}{%
    \begin{tabular}{p{3cm}ccc}
    \toprule
    \textbf{Subreddit} & \textbf{Creation Date} & \textbf{\#Raw Post} & \textbf{\#Filtered Post} \\ \midrule
    r/AIJobs & 24-Oct-17 & 135 & 9 \\ 
    r/Claude & 6-Dec-13 & 188 & 116 \\ 
    r/OpenSourceAI & 29-Jan-21 & 599 & 128 \\ 
    r/AITools & 24-Sep-20 & 639 & 78 \\ 
    r/NLP & 31-Dec-08 & 1,231 & 3 \\ 
    r/MistralAI & 13-Jun-23 & 1,397 & 380 \\ 
    r/PromptDesign & 10-Jun-22 & 1,617 & 411 \\ 
    r/HuggingFace & 29-Aug-21 & 1,663 & 349\\ 
    r/Anthropic & 4-Feb-23 & 2,058 & 808 \\ 
    r/ControlProblem & 29-Aug-15 & 3,434 & 203 \\ 
    r/LanguageTechnology & 10-Mar-10 & 4,177 & 895 \\ 
    r/AGI & 28-Jan-08 & 4,534 & 675 \\ 
    r/Automate & 26-Jun-12 & 5,182 & 415 \\ 
    r/PromptEngineering & 26-Feb-21 & 5,279 & 2,454 \\ 
    r/MLQuestions & 4-Mar-14 & 8,362 & 1,097 \\ 
    r/deepseek & 23-Nov-23 & 8,510 & 3,449 \\ 
    r/reinforcementlearning & 2-Mar-12 & 8,593 & 387 \\ 
    r/GPT3 & 17-Jul-20 & 9,078 & 1,978 \\ 
    r/deeplearning & 27-Nov-11 & 11,964 & 1,446 \\ 
    r/ChatGPTCoding & 6-Dec-22 & 14,802 & 5,333 \\ 
    r/compsci & 24-Mar-08 & 15,105 & 285 \\ 
    r/GitHub & 22-Oct-10 & 16,100 & 378 \\ 
    r/Bard & 6-Feb-09 & 19,141 & 6,905 \\ 
    r/learnmachinelearning & 23-Feb-16 & 37,431 & 4,434 \\ 
    r/Futurology & 12-Dec-11 & 41,684 & 893 \\ 
    r/artificial & 13-Mar-08 & 50,374 & 3,370 \\ 
    r/replika & 14-Mar-17 & 52,756 & 9,687 \\ 
    r/programming & 28-Feb-06 & 68,796 & 513 \\ 
    r/LocalLLaMA & 10-Mar-23 & 70,812 & 21,377 \\ 
    r/OpenAI & 11-Dec-15 & 73,004 & 18,262 \\ 
    r/ArtificialInteligence & 25-Feb-16 & 75,496 & 11,007 \\ 
    r/singularity & 28-Jan-08 & 77,729 & 9,111 \\ 
    r/MachineLearning & 29-Jul-09 & 82,513 & 5,689 \\ 
    r/technology & 25-Jan-08 & 128,322 & 13 \\ 
    r/webdev & 25-Jan-09 & 136,992 & 2,011 \\ 
    r/midjourney & 13-Mar-22 & 147,076 & 979 \\ 
    r/StableDiffusion & 23-Jul-22 & 162,138 & 3,012 \\ 
    r/Entrepreneur & 21-Aug-08 & 193,635 & 2,581 \\ 
    r/ChatGPT & 1-Dec-22 & 402,353 & 109,448 \\ 
    \bottomrule
    \end{tabular}}
     
    \caption{List of Subreddits Used for Data Collection}
    \label{tab:Subreddit_Overview}
\end{table}

\section{Dimensions of Trust and Distrust} \label{sec:app:dimensions-litearture}

Table \ref{tab :Initial List of Dimensions Found In Literature} provides the full list of dimensions extracted from the literature. This list was further refined for our study.

\begin{table}[ht]
\centering
    \resizebox{0.9\columnwidth}{!}{%
\begin{tabular}{lp{8cm}}
\toprule
\textbf{Dimensions}           & \textbf{Papers} \\ 
\midrule
Reliability          & \cite{zhang2023we, mcknight2009trust, bach2024systematic, braga2018survey, de2025mapping}   \\ 
Integrity            & \cite{mcknight2017distinguishing, harrison2001trust, mcknight2001while, braga2018survey, de2025mapping}   \\ 
Benevolence          & \cite{lankton2011does, lankton2015technology, zhang2024profiling, kim2024m, de2025mapping}   \\ 
Competence           & \cite{mcknight2017distinguishing, harrison2001trust, mcknight2001while, braga2018survey, xu2014different}   \\ 
Functionality        & \cite{xu2014different, lankton2011does, lankton2015technology}   \\ 
Security             & \cite{zhang2023we, bach2024systematic}   \\ 
Privacy              & \cite{zhang2023we}   \\ 
Transparency         & \cite{bach2024systematic, braga2018survey, klingbeil2024trust, shin2021effects, duenser2023trust, de2025mapping}   \\ 
Usability            & \cite{zhang2023we, xu2014different}   \\ 
Helpfulness          & \cite{mcknight2009trust, lankton2011does, lankton2015technology}   \\ 
Accuracy             & \cite{bach2024systematic, schuetz2025qualitative, salimzadeh2024dealing}   \\ 
Ethical use          & \cite{zhang2023we}   \\ 
Predictability       & \cite{harrison2001trust, mcknight2001while, braga2018survey}  \\ 
Responsiveness       & \cite{zhang2023we}   \\ 
Relatability         & \cite{bach2024systematic}   \\
Explainability       &  \cite{de2025mapping}   \\
Appearance           & \cite{xu2014different}   \\ 
Consistency          & \cite{mcknight2009trust}   \\ 
Decision-making      & \cite{salimzadeh2024dealing, klingbeil2024trust}   \\ 
Comprehension        & \cite{salimzadeh2024dealing}   \\ 
Appraisal            & \cite{salimzadeh2024dealing}   \\ 
Availability         & \cite{salimzadeh2024dealing}   \\ 
Relevance            & \cite{salimzadeh2024dealing}   \\ 
Interpretability     & \cite{salimzadeh2024dealing}   \\ 
Reliance             & \cite{klingbeil2024trust,salimzadeh2024dealing, zhang2024profiling}   \\ 
Algorithmic bias     & \cite{klingbeil2024trust}   \\ 
Contradiction        & \cite{klingbeil2024trust}   \\ 
Deception            & \cite{klingbeil2024trust}   \\ 
Complexity           & \cite{salimzadeh2024dealing}   \\ 
Difficulty           & \cite{salimzadeh2024dealing}   \\ 
Uncertainty          & \cite{salimzadeh2024dealing}   \\ 
Familiarity          & \cite{bach2024systematic}   \\ 
Malevolence          & \cite{zhang2024profiling,mcknight2017distinguishing}   \\ 
Deceit               & \cite{mcknight2017distinguishing}   \\ 
Incompetence         & \cite{mcknight2017distinguishing}   \\ 
\bottomrule
\end{tabular}}
\caption{Initial List of Dimensions In Literature}
\label{tab :Initial List of Dimensions Found In Literature}
\end{table}


\subsection{Two Approaches for Counting Dimensions}
As described in the Method section, we used two approaches to construct our ground-truth labels for dimensions of trust and distrust, based on the human annotation results. If more than half of the annotators tagged a dimension, it will be considered present in the post based on the `majority' approach; if any of the annotators tagged it, the dimension will be considered based on the `any' approach. Table \ref{tab:Dimensions} shows the overall number of posts tagged with each dimension in our annotated set using these two approaches.




\begin{figure*}[t]
    \centering

    \begin{subfigure}[t]{0.43\textwidth}
        \centering
        \includegraphics[width=\columnwidth]{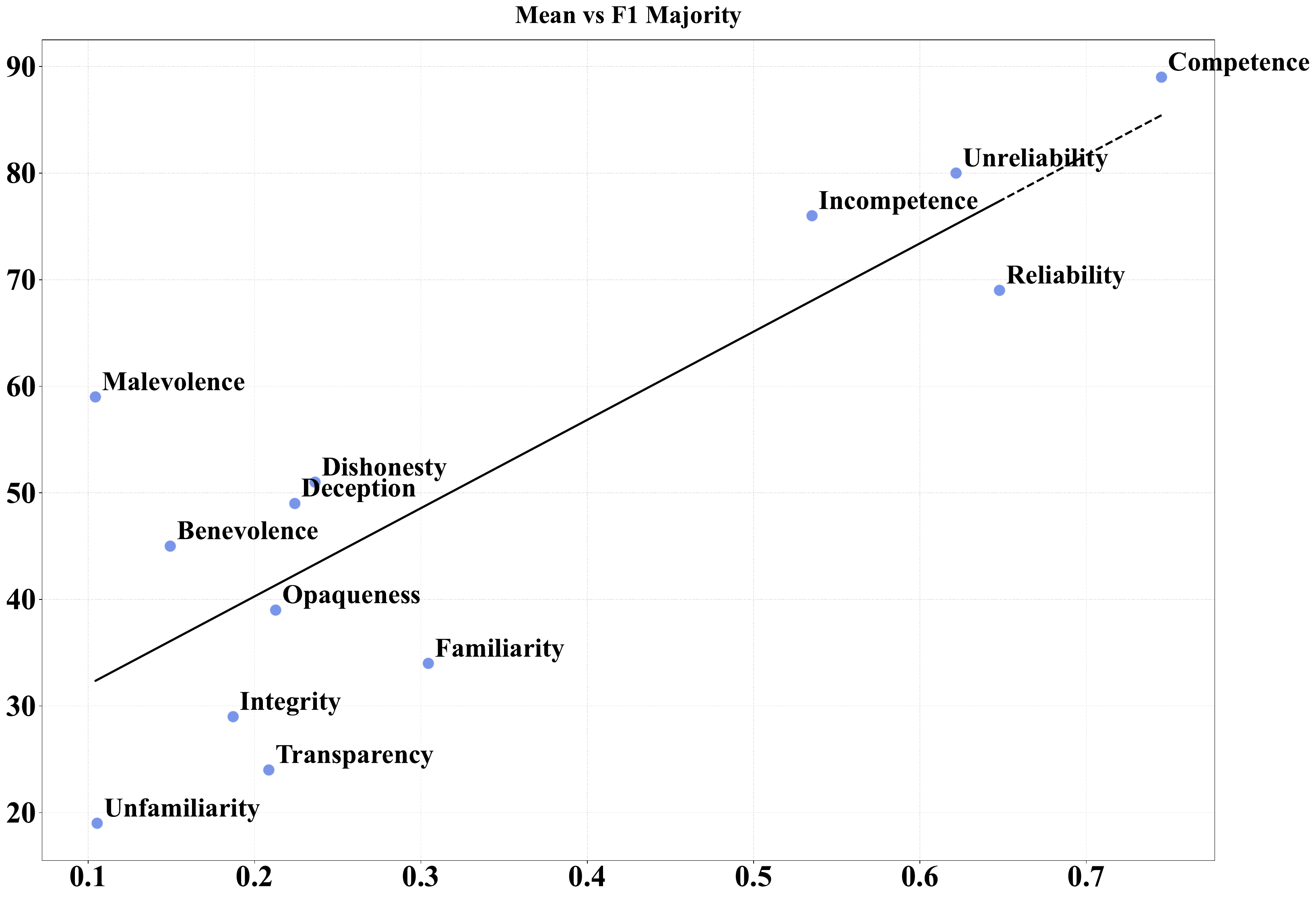}
        \caption{Majority approach ground-truth}
        \label{fig:classification_agreement_majority}
    \end{subfigure}
    \begin{subfigure}[t]{0.43\textwidth}
        \centering
        \includegraphics[width=\columnwidth]{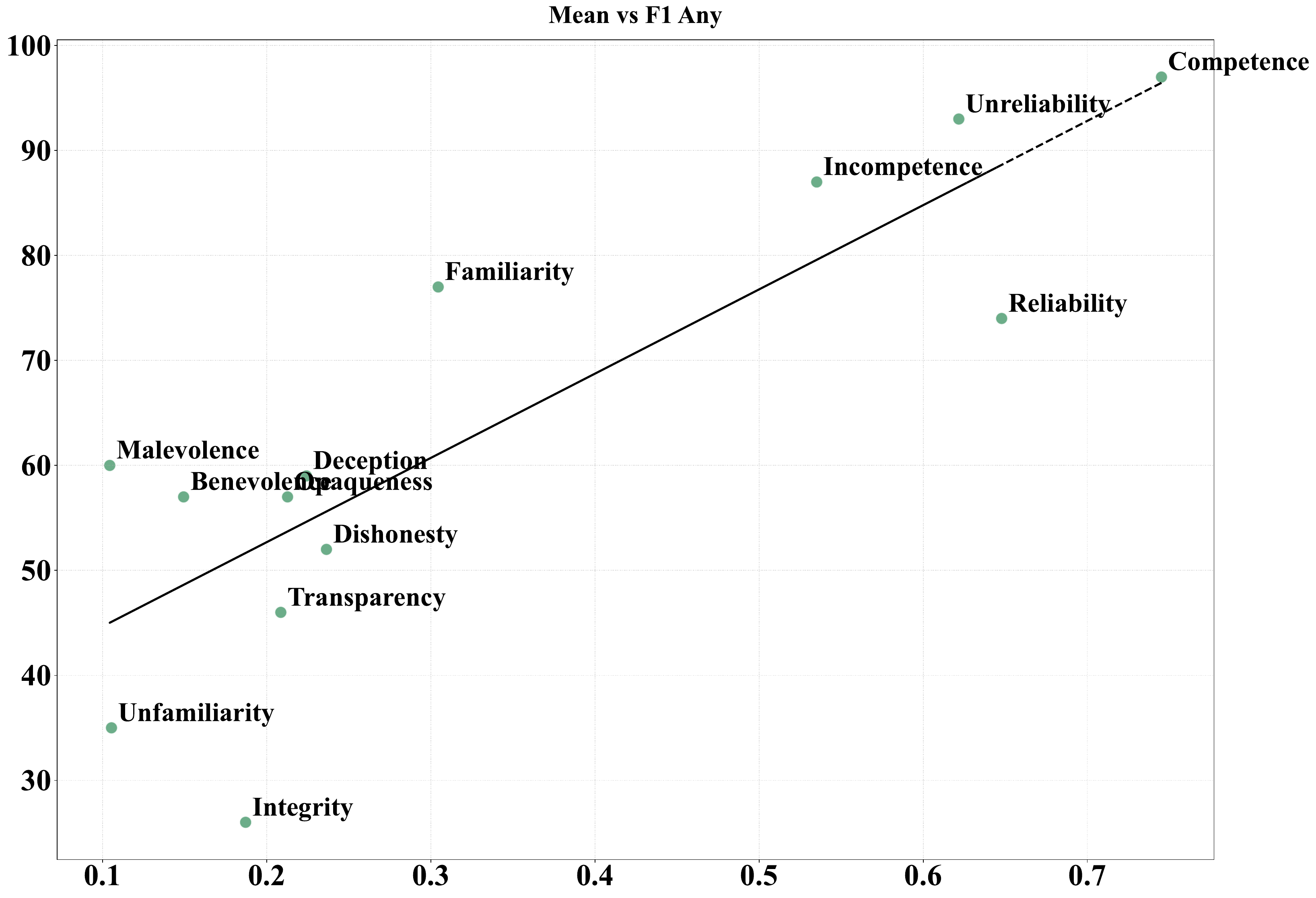}
        \caption{Any approach ground-truth}
        \label{fig:classification_agreement_any}
    \end{subfigure}
\vspace{-0.2cm}
    \caption{Relationship between human agreement rates and classification performance for dimensions}
    \label{fig:classification_agreement}
    \vspace{-0.2cm}
\end{figure*}

\section{Error Analysis} \label{sec:app:error-analysis}
Table \ref{tab:mistake-distribution} shows where misclassifications by LLM tend to happen using our test set. Most misclassifications occur where `Neither' is either wrongly predicted instead of `Trust' or `Distrust', or is wrongly predicted as one of them. While `Trust' is wrongly predicted as `Distrust' 8 times, the reverse has occurred only once, suggesting that the classifier is successful overall in not mistaking them for each other. 

\begin{table}[h]
\centering

\resizebox{0.8\columnwidth}{!}{
\begin{tabular}{@{}llc@{}}
\toprule
\textbf{Actual (Ground Truth)} & \textbf{Predicted Label} & \textbf{Count} \\ \midrule
Trust & Neither & 21             \\
Neither& Trust & 16             \\
Distrust & Neither & 14             \\
Neither & Distrust & 11             \\
Trust & Distrust  & 8              \\
Both  & Distrust  & 7              \\
Both & Trust & 5              \\
Trust  & Both & 2              \\
Distrust & Trust & 1              \\ \bottomrule
\end{tabular}

}
\caption{\footnotesize Common Mistakes With the Best Model}
\label{tab:mistake-distribution}
\end{table}

\section{Annotation Procedure}\label{sec:app-annot-process}

\subsection{Annotation Sample}\label{sec:crowdsourcing details}
We draw a sample of 3K posts from our data for crowdsource annotation. However, we intentionally avoided complete random sampling. Given that the focus of this study is primarily on trust and distrust, we first ran two open-source LLM models (Mixtral and Llama) on 5K posts and kept those where two models agreed on either of these labels: `Trust', `Distrust', `Both'. We then collected a random sample from each class in a way that the final sample includes 40 \% trust, 40 \% distrust, and 20 \% neither in the annotation sample. We took this step to ensure most annotated samples are related to trust or distrust. This explains the difference in the distribution of labels between the sample set and the entire collected data.

\subsection{Annotation Guideline and Questions}
\label{sec:app-annot-guide}

The annotators were given the following guidelines and questions:

{\renewcommand{\lstlistingname}{Annotation}%
\begin{lstlisting}[caption={ Guideline.},xleftmargin=0pt,xrightmargin=0pt]
@\textcolor{darkblue}{Task Introduction}@
The purpose of this annotation task is to understand how people discuss trust and distrust toward Generative AI (GenAI) and the companies responsible for its development: 
- Trust is defined as a belief in the reliability, competence, or integrity of GenAI technologies / responsible companies, leading to positive expectations about their performance and ethical behavior.  
- Distrust is defined as skepticism or concern about the reliability, competence, or ethical implications of GenAI / responsible companies, leading to negative expectations or cautious behavior toward these technologies or companies. 
You will identify the presence or absence of trust or distrust in each text, as well as the entities who are trustors (those who express trust or distrust) of specific GenAI systems/ companies. Additionally, you will annotate the dimensions of trust or distrust, which include aspects such as reliability, transparency, integrity, and deception.  
@\textcolor{darkblue}{Instruction}@
Please note that you need to answer all the necessary questions for each text to move forward to the next one, otherwise you will see an error that keeps you from moving to the next text.  
This instruction is accessible throughout the annotation through the ``Help'' button on the upper-left side of the screen. Please make sure to use it at any time you need to review the definitions and instructions. 
@\textcolor{darkblue}{Identify Trust or Distrust}@
For each text, determine whether it primarily expresses:  
Trust 
Distrust  
Both  
Neither  
If the label is Neither, move on to the next post by skipping the following questions. If the label is Trust, Distrust, or Both, continue.  
@\textcolor{darkblue}{Label Trust Dimensions}@
Here are the definitions of trust dimensions. Choose one or more that are explicitly or implicitly mentioned:  
- Reliability (Consistent and accurate performance): Consistency and accuracy of the information provided by LLMs.
- Transparency (Openness about how AI works): Being open and clear about how data is collected, processed, and how LLM works.  
- Familiarity (Previous experience with AI): Previous experience with AI or tools that makes LLMs more familiar.
- Integrity (Ethical use of AI): AI tool following a set of (ethical) principles that the user finds acceptable. 
- Competence (Ability to perform tasks effectively): Having the capability, functionality, or features to do what users want
- Benevolence (Good intentions behind AI): Belief that LLMs want to do good to the user and have good intentions. 
@\textcolor{darkblue}{Label Distrust Dimensions}@
Here are the definitions of distrust dimensions. Choose one or more that are explicitly or implicitly mentioned:  
- Unreliability (Inconsistent, inaccurate results): Consistency and accuracy of the information provided by LLMs.
- Opaqueness (Lack of openness or explainability): Being open and clear about how data is collected, processed, and how LLM works. 
- Unfamiliarity (Uncertainty due to new or unfamiliar tech): Previous experience with AI or tools that makes LLMs more familiar.
- Dishonesty (Ethical concerns about lack of integrity, bias, manipulation): AI tool following a set of (ethical) principles that the user finds acceptable.
- Incompetence (Fails to perform tasks effectively): Having the capability, functionality, or features to do what users want
- Deception (Misleading or fabricated responses): Belief that LLM is dishonest and potentially provides false information, e.g., hallucination and make up stuff.
- Malevolence (Harmful or dangerous AI use): Belief that LLMs have the intention to harm the user, don't want to do good to the user.
@\textcolor{darkblue}{Identify Reasons}@
What reasons for trust or distrust are mentioned in the text? Identify the options from the list provided in the questions.  
@\textcolor{darkblue}{Identify Trustor}@
Trustor:The entity expressing trust or distrust.  
Available options are:  
- Generative AI users
- Software developers
- Researchers or academics
- Tech industry professionals
- General public 
- Media and journalists 
- Business leaders or executives
- AI ethicists or advocacy groups
- Artists or creatives 
- Educators and knowledge workers
- Other  
@\textcolor{darkblue}{Additional Points}@
This annotation task focuses on Generative AI and large language models (LLMs) such as ChatGPT, and companies responsible for their development (e.g., OpenAI). If trust is directed toward unrelated factors (e.g., hardware issues, people, or unrelated software), those should not be annotated. Focus only on trust or distrust related to LLMs, GenAI technologies, and responsible companies.  
\end{lstlisting}}

{\renewcommand{\lstlistingname}{Annotation}%
\begin{lstlisting}[caption={Main Questions.},xleftmargin=0pt,xrightmargin=0pt]

@\textcolor{darkblue}{Question 1: }@ Does the post express any level of trust or distrust in generative AI models or companies (e.g., ChatGPT, Llama, Open AI, or similar)?  
- Trust
- Distrust
- Both Trust and Distrust
- Neither, No trust or distrust
@\textcolor{darkblue}{Question 2: }@
Which dimensions of trust are discussed or implied in the post? (Select all that apply)  
Reliability, Transparency, Familiarity, Integrity, Competence,  Benevolence, None of the above.  
@\textcolor{darkblue}{Question 3: }@
Which dimensions of distrust are discussed or implied in the post? (Select all that apply)  
Unreliability, Opaqueness, Unfamiliarity, Dishonesty, Incompetence,  Deception, Malevolence, None of the above.  
@\textcolor{darkblue}{Question 4: }@
What is the main reason for trust or distrust expressed in the post? (Select one primary reason)  
- Personal experience using AI models
- Reports from media or experts
- Discussions on social media or online forums
- General perception of AI technology
- General perception toward responsible companies
- Experiences from friends, families, colleagues, etc. who used it
- Other  
@\textcolor{darkblue}{Question 5: }@
Who is expressing trust or distrust? (Trustor)  
- Generative AI users
- Software developers
- Researchers or academics  
- Tech industry professionals 
- General public
- Media and journalists 
- Business leaders or executives  
- AI ethicists or advocacy groups
- Artists or creatives
- Educators and knowledge workers
- Other  
\end{lstlisting}}

{\renewcommand{\lstlistingname}{Annotation}%
\begin{lstlisting}[caption={Demographic Questions.},xleftmargin=0pt,xrightmargin=0pt]
@\textcolor{darkblue}{How old are you?}@ 
- 18-24 
- 25-34 
- 35-44 
- 45-54 
- 55-64 
- 65 or older 
- Prefer not to say 
@\textcolor{darkblue}{Which of the following genders do you most identify with?}@ 
- Male 
- Female 
- Non-binary/ third gender 
- Others 
- Prefer not to answer 
@\textcolor{darkblue}{Which of the following best describes you? Select all that apply.}@ 
- Asian 
- Black or African American 
- Hispanic or Latino 
- Native American or Alaska Native 
- Native Hawaiian or Pacific Islander 
- White 
- Other 
@\textcolor{darkblue}{What is the highest level of education you have completed?}@  
- Some high school, no diploma 
- High school diploma or GED 
- Some college, no degree 
- Associate (2-year) degree 
- Bachelor's (4-year) degree 
- Master's degree 
- Doctorate degree 
- Prefer not to say 
@\textcolor{darkblue}{Which of the following best describes your employment status?}@ 
- Employed full-time 
- Employed part-time 
- Student 
- Disabled 
- Retired 
- Unemployed 
- Prefer not to say 
@\textcolor{darkblue}{Which of the following best describes your total annual income?}@ 
- Under 30,000 
- 30,000 to 49,999 
- 50,000 to 74,999 
- 75,000 to 99,999 
- 100,000 to 149,999 
- 150,000 or more 
- Prefer not to say 
@\textcolor{darkblue}{Which political party do you typically support?}@ 
- Democratic Party 
- Republican Party 
- Libertarian Party 
- Green Party 
- Independent 
- Other 
- Prefer not to say 
@\textcolor{darkblue}{What is your native language?}@
- English  
- Spanish  
- Arabic  
- Chinese  
- Hindi  
- Other 
- Prefer not to say 
@\textcolor{darkblue}{How would you describe your level of comfort and proficiency with technology?}@ 
- Beginner: I have limited experience with technology and often need assistance. 
- Intermediate: I am comfortable using common technology and can troubleshoot basic issues. 
- Advanced: I am very comfortable with technology and can handle complex software or troubleshoot issues independently. 
- Expert: I have extensive knowledge of technology, including programming, system configuration, and troubleshooting complex issues
- Prefer not to say 
@\textcolor{darkblue}{How familiar are you with large language models (LLMs) like ChatGPT, Llama, or others?}@ 
- Not familiar: I have never heard of or used large language models. 
- Somewhat familiar: I've heard of large language models and know a little about what they do. 
- Moderately familiar: I understand the basic functions of large language models and have some experience with them. 
- Very familiar: I have a strong understanding of large language models and regularly use or study them. 
- Expert: I have extensive knowledge and experience with large language models, possibly in a professional or research capacity. 
- Prefer not to say 
@\textcolor{darkblue}{How often do you use LLMsbased models or chatbots?}@
- Never 
- Rarely 
- Occasionally 
- Frequently 
- Always 
- Prefer not to say 
@\textcolor{darkblue}{If you are familiar with large language models e.g. ChatGPT, how would you describe your level of trust in them?}@
- I fully trust them  
- I somewhat trust them 
- I'm not sure 
- I somewhat distrust them 
- I fully distrust them 
- Prefer not to say 

\end{lstlisting}}

\subsection{Annotation Interface and Procedure}
\label{sec:app-annotation-procedure}
The study interface included five components: (1) an introduction page describing the study purpose, (2) an instruction page with detailed guidelines and definitions, (3) a question page with five items per post, (4) a demographic page with 12 questions, and (5) an experience page for feedback. The session concluded with an end page that submitted responses and redirected participants to Prolific. A Help button allowed access to instructions at any point. We piloted the interface in-house (3 participants, 50 posts each) and ran two Potato tests (5 and 20 participants) before deploying the main study. 
Each round recruited 100 Prolific participants to annotate 50 posts each, yielding 1,000 posts per round with five annotations per post. Three attention checks were embedded, and only responses from participants who passed all were retained. To replace missing data, three supplemental rounds (17, 17, and 12 participants) were added. Eligibility required U.S.-based English speakers with a 95–100\% approval rate and 100–10,000 prior submissions. Participants were paid above minimum wage, with the research team available for support throughout.

\section{Prompts Used for Classification Tasks}\label{sec:app-prompts}
This section provides the prompts used in each classification task. 

\setcounter{lstlisting}{0}
{\renewcommand{\lstlistingname}{Prompt}%
\begin{lstlisting}[caption={Trust vs. Distrust},xleftmargin=0pt,xrightmargin=0pt]
@\textcolor{darkblue}{Task Introduction}@ 
You are an advanced AI model trained for text classification. The purpose of this annotation task is to classify how people discuss trust and distrust toward Generative AI dimensions of trust in text. We define trust as a belief in the reliability, competence, or integrity of GenAI technologies, leading to positive expectations about their performance and ethical behavior. On the other hand, we define distrust as skepticism or concern about the reliability, competence, or ethical implications of GenAI, leading to negative expectations or cautious behavior toward these technologies. 
You will identify the presence or absence of trust or distrust in each text, as well as the dimensions of trust or distrust, which include aspects such as reliability, transparency, integrity and deception. 
Your task is to classify a given text:  
Classify the text into a high-level category based on showing trust or distrust toward GenAI.Follow the label definitions strictly.  

Classify the text into one of the following categories in terms of showing trust or distrust toward Generative AI:  
- Trust: The text expresses belief in the reliability, competence, or integrity of GenAI technologies, leading to positive expectations about their performance and ethical behavior.
- Distrust: The text expresses skepticism or concern about the reliability, competence, or ethical implications of GenAI, leading to negative expectations or cautious behavior toward these technologies or companies.
- Both: The text indicates both trust or distrust in some way. Such a text shows tendencies that suggest both trust and distrust at the same time.
- Neither: The text is not expressing trust or distrust towards GenAI, or is irrelevant to the context of our study.
@\textcolor{darkblue}{Output Format}@ 
[Label]  
Strictly follow this format and don't add any other text.
Read the following text and provide the best labels: {text} 
\end{lstlisting}}

{\renewcommand{\lstlistingname}{Prompt}%
\begin{lstlisting}[caption={Dimensions of Trust vs. Distrust},xleftmargin=0pt,xrightmargin=0pt]
@\textcolor{darkblue}{Task Instruction}@ 
You are an advanced AI model trained for text classification..
We define trust as a belief in the reliability, competence, or integrity of GenAI technologies, leading to positive expectations about their performance and ethical behavior.
On the other hand, we define distrust as skepticism or concern about the reliability, competence, or ethical implications of GenAI, leading to negative expectations or cautious behavior toward these technologies or companies.
Identify the dimensions of trust/distrust toward GenAI in text based on {task1_label}.
@\textcolor{darkblue}{Conditional instructions}@
if task1_label = Trust:
    Choose one or more of the following dimensions: Reliability, Transparency, Familiarity, Integrity, Competence, Benevolence
elif task1_label = Distrust:
    Choose one or more of the following dimensions: Unreliability, Opaqueness, Unfamiliarity, Dishonesty, Incompetence, Deception, Malevolence
elif task1_label = Both:
    Choose one or more of the dimensions from both lists
else:  (Neither)
    No further classification is required. Simply output Neither.
@\textcolor{darkblue}{Output format instructions}@
After labeling the text with dimensions, add a brief cue that explains why you chose each of the dimensions.
Use the following format for your response:
Dimensions: [Dimension 1, Dimension 2, ...]
Cues: [Cue 1, Cue 2, ...]
Strictly follow this format and don't add any other text.
Read the following text and provide the best dimensions: {text}
\end{lstlisting}}

{\renewcommand{\lstlistingname}{Prompt}%
\begin{lstlisting}[caption={Detection of Trustor},xleftmargin=0pt,xrightmargin=0pt]
@\textcolor{darkblue}{Task Instruction}@ 
You are an assistant in detecting the trustor of trust or distrust towards Generative AI or Large Language Models mentioned in the text.  
We define trust as a belief in the reliability, competence, or integrity of GenAI technologies/responsible companies, leading to positive expectations about their performance and ethical behavior. On the other hand, we define distrust as skepticism or concern about the reliability, competence, or ethical implications of GenAI/responsible companies, leading to negative expectations or cautious behavior toward these technologies or companies.  
Trustor: The entity expressing trust or distrust. Trustors could be users of GenAI, software developers, or others.  
You will identify the trustor of trust or distrust in each text from the following options:
- Generative AI users
- Software developers
- Researchers or academics
- Tech industry professionals
- General public
- Media and journalists
- Business leaders or executives
- AI ethicists or advocacy groups
- Artists or creatives
- Educators and knowledge workers
- Other
@\textcolor{darkblue}{Output format:}@  
[Label ID ONLY e.g. 1-7]
Do not add any extra text  

Here is the text: {text} 
\end{lstlisting}}

{\renewcommand{\lstlistingname}{Prompt}%
\begin{lstlisting}[caption={Detection of Reasons},xleftmargin=0pt,xrightmargin=0pt]
@\textcolor{darkblue}{Task Instruction}@ 
You are an assistant in detecting the reasons for trust or distrust towards Generative AI or Large Language Models mentioned in the text.  
We define trust as a belief in the reliability, competence, or integrity of GenAI technologies/responsible companies, leading to positive expectations about their performance and ethical behavior. On the other hand, we define distrust as skepticism or concern about the reliability, competence, or ethical implications of GenAI/responsible companies, leading to negative expectations or cautious behavior toward these technologies or companies.  
You will identify the reason of trust or distrust in each text from the following options:
- Personal experience using AI models
- Reports from media or experts
- Discussions on social media or online forums
- General perception of AI technology
- General perception toward responsible companies
- Experiences from friends, families, colleagues, etc.
- Other
@\textcolor{darkblue}{Output format:}@  
[Label ID ONLY e.g. 1-7]
Do not add any extra text.

Here is the text: {text} 
\end{lstlisting}}

\section{Omitted Details from Section~\ref{sec:result}}
\label{sec:appendix-results}

\begin{table}[ht!]
\small
\centering
\resizebox{0.8\columnwidth}{!}{%
\begin{tabular}{lllrr}
\toprule
\textbf{} & \textbf{dim\_1} & \textbf{dim\_2} & \textbf{co\_count} & \textbf{Jaccard} \\
\midrule
\multirow{10}{*}{\rotatebox{90}{Trust}} 
 & Competence   & Familiarity     & 55,275  & \textbf{0.76} \\
 & Competence   & Reliability     & 47,191  & 0.66 \\
 & Familiarity  & Reliability     & 36,061  & 0.52 \\
 & Competence   & Transparency    & 34,206  & 0.47 \\
  & Familiarity  & Transparency    & 27,765  & 0.43 \\
 & Transparency & Reliability     & 24,614  & 0.42 \\
 & Competence   & Benevolence     & 20,546  & 0.29 \\
  & Familiarity  & Benevolence     & 16,071   & 0.25 \\
 & Reliability    & Benevolence     & 13,695   & 0.24 \\
 & Transparency & Benevolence       & 8,481   & 0.17 \\

\midrule
\multirow{10}{*}{\rotatebox{90}{Distrust}} 
 & Incompetence & Unreliability   & 39416  & \textbf{0.67} \\
 & Opaqueness   & Unreliability   & 21518  & 0.36 \\
  & Deception    & Unreliability   & 20967   & 0.36 \\
  & Unreliability & Unfamiliarity  & 20800    & 0.35 \\
& Unreliability   & Dishonesty      & 14744   & 0.25 \\
& Incompetence & Deception      &  13374  &  0.26 \\
& Incompetence & Unfamiliarity   &  13084 &  0.24  \\
& Unfamiliarity& Opaqueness      & 12837   & 0.36 \\
& Opaqueness   & Incompetence    & 12327  & 0.23 \\
 & Dishonesty   & Deception       & 10770   & 0.35 \\
\bottomrule
\end{tabular}}
\caption{Most Co-occurrent Pairs of Dimensions in Trust and Distrust}
\label{tab:Most Co-occurrent Pairs of Dimensions}
\end{table}

\begin{table}[h]
\centering
\resizebox{0.5\columnwidth}{!}{
\begin{tabular}{lc}
\hline
\textbf{Dimension} & \textbf{Aggregate} \\ \hline
Competence & 0.75 \\
Familiarity & 0.30 \\
Integrity & 0.19 \\
Reliability & 0.65 \\
Transparency & 0.21 \\
Benevolence & 0.15 \\
Deception & 0.22 \\
Dishonesty & 0.24 \\
Incompetence & 0.54 \\
Malevolence & 0.10 \\
Opaqueness & 0.21 \\
Unreliability & 0.62 \\
Unfamiliarity & 0.11 \\ \hline
\end{tabular}}
\caption{Aggregate Ratios of Dimensions Tagged by Human Annotators}
\label{tab:dimension_means_small}
\end{table}


\begin{table*}[t]
\centering
\resizebox{0.7\textwidth}{!}{%
\begin{tabular}{cllllllllllllllcc}
\multicolumn{1}{l}{Type} &
  Model &
  \multicolumn{13}{c}{Dimensions} &
  \multicolumn{2}{c}{Average} \\
  \hline
\multicolumn{1}{l}{} &
   &
  Ben &
  Com &
  Dec &
  Dis &
  Fam &
  Inc &
  Int &
  Mal &
  Opa &
  Rel &
  Tra &
  Unf &
  Unr &
  \multicolumn{1}{l}{\begin{tabular}[c]{@{}l@{}}Avg\\F1\end{tabular}} \\
  \hline

\multirow{13}{*}{Zero-Shot} &
  gpt-oss-20b &
  0.35 & 0.84 & 0.48 & 0.48 & 0.25 & 0.74 & 0.28 & 0.52 & 0.35 & 0.59 & 0.18 & 0.09 & 0.77 & 0.46 \\
&
  Mixtral-8x7B &
  0.24 & 0.87 & 0.35 & 0.36 & 0.32 & 0.48 & 0.19 & \textbf{0.59} & 0.30 & 0.54 & 0.18 & 0.10 & 0.77 & 0.41 \\
&
  Llama-4 &
  0.36 & \textbf{0.89} & 0.45 & 0.41 & 0.25 & 0.75 & 0.22 & 0.51 & 0.27 & 0.43 & 0.19 & 0.14 & 0.74 & 0.43 \\
&
  gpt-5.2-2025-12-11 &
  0.38 & 0.87 & 0.44 & 0.39 & 0.28 & 0.72 & 0.21 & 0.37 & 0.27 & 0.53 & 0.15 & \textbf{0.19} & 0.78 & 0.43 \\
&
  gpt-5.1-2025-11-13 &
  0.34 & 0.88 & 0.44 & 0.39 & 0.29 & 0.71 & 0.17 & 0.41 & 0.32 & 0.60 & 0.10 & 0.16 & 0.73 & 0.43 \\
&
  gpt-5-2025-08-07 &
  0.32 & 0.88 & 0.43 & 0.42 & 0.32 & 0.74 & \textbf{0.29} & 0.52 & 0.28 & 0.42 & 0.20 & 0.11 & 0.67 & 0.43 \\
&
  gpt-5-mini-2025-08-07 &
  0.36 & \textbf{0.89} & 0.42 & 0.45 & 0.31 & \textbf{0.76} & 0.25 & 0.55 & 0.34 & 0.68 & 0.20 & 0.11 & 0.79 & 0.47 \\
&
  gpt-5-nano-2025-08-07 &
  0.33 & 0.87 & 0.46 & 0.40 & 0.29 & 0.75 & 0.19 & 0.46 & 0.36 & 0.58 & 0.22 & 0.11 & 0.76 & 0.44 \\
&
  gpt-4o-2024-11-20 &
  0.35 & \textbf{0.89} & 0.45 & 0.39 & 0.30 & 0.73 & 0.20 & 0.43 & \textbf{0.39} & 0.46 & 0.14 & 0.15 & 0.77 & 0.43 \\
&
  gpt-4o-mini-2024-07-18 &
  0.37 & 0.87 & 0.41 & 0.48 & 0.27 & 0.73 & 0.21 & 0.49 & 0.34 & 0.55 & 0.16 & 0.11 & 0.79 & 0.44 \\
&
  gpt-4.1-2025-04-14 &
  0.41 & 0.88 & 0.48 & 0.42 & 0.31 & 0.75 & 0.26 & 0.43 & 0.37 & 0.60 & 0.14 & 0.12 & 0.78 & 0.46 \\
&
  gpt-4.1-mini-2025-04-14 &
  0.31 & 0.88 & 0.47 & 0.35 & 0.32 & 0.70 & 0.23 & 0.41 & 0.32 & 0.52 & 0.10 & 0.10 & 0.77 & 0.42 \\
&
  gpt-4.1-nano-2025-04-14 &
  0.31 & 0.81 & 0.36 & 0.29 & 0.31 & 0.48 & 0.18 & 0.54 & 0.25 & 0.59 & 0.20 & 0.07 & 0.75 & 0.39 \\
  \hline

\multirow{13}{*}{Few-Shot} &
  gpt-oss-20b &
  0.38 & 0.88 & 0.46 & 0.35 & 0.28 & 0.69 & 0.24 & 0.54 & 0.34 & 0.63 & 0.22 & 0.09 & 0.76 & 0.45 \\
&
  Mixtral-8x7B &
  0.20 & 0.88 & 0.25 & 0.41 & \textbf{0.34} & 0.66 & 0.17 & 0.45 & 0.27 & 0.59 & 0.21 & 0.12 & 0.78 & 0.41 \\
&
  Llama-4 &
  0.37 & \textbf{0.89} & 0.42 & 0.38 & 0.22 & 0.72 & 0.22 & 0.43 & 0.29 & 0.48 & 0.19 & 0.04 & 0.72 & 0.41 \\
&
  gpt-5.2-2025-12-11 &
  0.41 & 0.87 & 0.45 & 0.35 & 0.27 & 0.74 & 0.20 & 0.37 & 0.32 & 0.53 & 0.16 & 0.11 & 0.78 & 0.43 \\
&
  gpt-5.1-2025-11-13 &
  0.38 & 0.88 & 0.47 & 0.36 & 0.32 & 0.71 & 0.19 & 0.41 & 0.32 & \textbf{0.69} & \textbf{0.24} & 0.17 & 0.77 & 0.45 \\
&
  gpt-5-2025-08-07 &
  0.29 & \textbf{0.89} & 0.46 & 0.38 & \textbf{0.34} & 0.75 & 0.27 & 0.54 & 0.31 & 0.50 & 0.22 & 0.15 & 0.74 & 0.45 \\
&
  gpt-5-mini-2025-08-07 &
  0.36 & \textbf{0.89} & 0.45 & 0.42 & \textbf{0.34} & 0.73 & 0.25 & 0.51 & 0.34 & 0.57 & 0.19 & 0.15 & \textbf{0.80} & 0.46 \\
&
  gpt-5-nano-2025-08-07 &
  0.29 & 0.86 & 0.42 & 0.27 & 0.32 & 0.70 & 0.21 & 0.53 & 0.32 & 0.62 & 0.19 & 0.09 & 0.77 & 0.43 \\
&
  gpt-4o-2024-11-20 &
  0.32 & 0.88 & 0.45 & 0.39 & 0.33 & 0.74 & 0.25 & 0.42 & 0.37 & 0.44 & 0.20 & 0.11 & 0.78 & 0.44\\
&
  gpt-4o-mini-2024-07-18 &
  \textbf{0.45} & 0.88 & 0.44 & \textbf{0.51} & 0.30 & 0.73 & 0.20 & 0.44 & 0.35 & 0.57 & 0.15 & 0.13 & 0.78 & 0.46 \\
&
  gpt-4.1-2025-04-14 &
  0.43 & 0.88 & \textbf{0.49} & 0.37 & \textbf{0.34} & 0.75 & 0.23 & 0.44 & 0.38 & 0.67 & 0.20 & 0.12 & \textbf{0.80} & 0.47 \\
&
  gpt-4.1-mini-2025-04-14 &
  0.30 & 0.86 & 0.45 & 0.34 & 0.32 & 0.67 & 0.23 & 0.35 & 0.35 & 0.61 & 0.17 & \textbf{0.19} & 0.75 & 0.43 \\
&
  gpt-4.1-nano-2025-04-14 &
  0.31 & 0.78 & 0.41 & 0.22 & 0.33 & 0.60 & 0.18 & 0.48 & 0.28 & 0.65 & 0.17 & 0.07 & 0.76 & 0.40 \\

\bottomrule
\end{tabular}}
\caption{Classification of dimensions of trust and distrust by zero-shot and few-shot models, and the scores used for our criteria under the Majority Criteria. Dimensions: Ben: Benevolence, Com: Competence, Dec: Deception, Dis: Dishonesty, Fam: Familiarity, Inc: Incompetence, Int: Integrity, Mal: Malevolence, Opa: Opaqueness, Rel: Reliability, Tra: Transparency, Unf: Unfamiliarity, Unr: Unreliability.
}
\label{tab:majority_task_style}
\end{table*}

\begin{table*}[t]
\centering
\resizebox{0.7\textwidth}{!}{%
\begin{tabular}{cllllllllllllllcc}
\multicolumn{1}{l}{Type} &
  Model &
  \multicolumn{13}{c}{Dimensions} &
  \multicolumn{2}{c}{Average} \\
  \hline
\multicolumn{1}{l}{} &
   &
  Ben &
  Com &
  Dec &
  Dis &
  Fam &
  Inc &
  Int &
  Mal &
  Opa &
  Rel &
  Tra &
  Unf &
  Unr &
  \multicolumn{1}{l}{\begin{tabular}[c]{@{}l@{}}Avg\\F1\end{tabular}} \\
  \hline

\multirow{13}{*}{Zero-Shot} &
  gpt-oss-20b &
  0.35 & 0.90 & 0.49 & 0.40 & 0.43 & 0.84 & 0.12 & 0.33 & 0.36 & 0.59 & 0.11 & 0.18 & 0.81 & 0.46 \\
&
  Mixtral-8x7B &
  0.47 & 0.95 & 0.35 & 0.45 & 0.70 & 0.48 & 0.12 & 0.33 & 0.53 & 0.56 & 0.25 & 0.15 & \textbf{0.93} & 0.48 \\
&
  Llama-4 &
  0.32 & \textbf{0.97} & 0.56 & 0.42 & 0.30 & 0.86 & 0.23 & 0.42 & 0.20 & 0.41 & 0.11 & 0.09 & 0.75 & 0.43 \\
&
  gpt-5.2-2025-12-11 &
  0.27 & 0.94 & 0.48 & 0.47 & 0.39 & 0.77 & 0.15 & 0.55 & 0.20 & 0.52 & 0.11 & 0.09 & 0.79 & 0.44 \\
&
  gpt-5.1-2025-11-13 &
  0.47 & 0.95 & 0.51 & 0.29 & 0.75 & 0.68 & 0.07 & 0.49 & 0.22 & 0.59 & 0.04 & 0.13 & 0.74 & 0.46 \\
&
  gpt-5-2025-08-07 &
  0.24 & 0.96 & 0.41 & 0.47 & 0.57 & \textbf{0.87} & 0.14 & 0.32 & 0.35 & 0.39 & 0.15 & 0.17 & 0.68 & 0.44 \\
&
  gpt-5-mini-2025-08-07 &
  0.41 & \textbf{0.97} & 0.53 & \textbf{0.52} & 0.73 & \textbf{0.87} & 0.20 & 0.36 & 0.52 & 0.68 & 0.23 & 0.23 & 0.84 & 0.55 \\
&
  gpt-5-nano-2025-08-07 &
  0.47 & 0.94 & 0.42 & 0.28 & 0.66 & 0.82 & 0.17 & 0.23 & 0.33 & 0.58 & 0.16 & 0.16 & 0.81 & 0.46 \\
&
  gpt-4o-2024-11-20 &
  0.43 & \textbf{0.97} & \textbf{0.59} & 0.45 & 0.75 & 0.77 & 0.10 & 0.53 & 0.44 & 0.44 & 0.13 & 0.20 & 0.82 & 0.51 \\
&
  gpt-4o-mini-2024-07-18 &
  0.29 & 0.94 & 0.40 & 0.38 & 0.48 & 0.83 & 0.09 & 0.42 & 0.50 & 0.53 & 0.11 & 0.29 & 0.91 & 0.48 \\
&
  gpt-4.1-2025-04-14 &
  0.29 & 0.96 & 0.52 & 0.36 & \textbf{0.77} & 0.81 & 0.10 & 0.49 & 0.48 & 0.58 & 0.09 & 0.30 & 0.85 & 0.51 \\
&
  gpt-4.1-mini-2025-04-14 &
  0.49 & 0.96 & 0.50 & 0.23 & 0.64 & 0.74 & 0.07 & 0.52 & 0.43 & 0.51 & 0.08 & 0.20 & 0.80 & 0.47 \\
&
  gpt-4.1-nano-2025-04-14 &
  0.42 & 0.86 & 0.28 & 0.18 & 0.60 & 0.41 & 0.07 & 0.36 & 0.48 & 0.60 & 0.26 & \textbf{0.35} & 0.81 & 0.44 \\
  \hline

\multirow{13}{*}{Few-Shot} &
  gpt-oss-20b &
  0.43 & 0.96 & 0.45 & 0.29 & 0.42 & 0.68 & 0.17 & 0.41 & 0.43 & 0.62 & 0.17 & 0.09 & 0.77 & 0.45 \\
&
  Mixtral-8x7B &
  \textbf{0.57} & 0.95 & 0.22 & 0.51 & 0.68 & 0.71 & \textbf{0.26} & 0.46 & 0.55 & 0.66 & 0.35 & 0.23 & 0.90 & 0.54 \\
&
  Llama-4 &
  0.39 & \textbf{0.97} & 0.54 & 0.41 & 0.22 & 0.76 & 0.22 & 0.46 & 0.19 & 0.44 & 0.21 & 0.10 & 0.73 & 0.43 \\
&
  gpt-5.2-2025-12-11 &
  0.37 & 0.93 & 0.44 & 0.41 & 0.41 & 0.73 & 0.17 & 0.53 & 0.25 & 0.51 & 0.18 & 0.07 & 0.77 & 0.44 \\
&
  gpt-5.1-2025-11-13 &
  0.44 & 0.96 & 0.48 & 0.32 & 0.67 & 0.67 & 0.13 & 0.50 & 0.26 & 0.72 & 0.12 & 0.14 & 0.75 & 0.48 \\
&
  gpt-5-2025-08-07 &
  0.26 & \textbf{0.97} & 0.40 & 0.41 & 0.58 & 0.78 & 0.16 & 0.42 & 0.45 & 0.47 & 0.21 & 0.18 & 0.74 & 0.46 \\
&
  gpt-5-mini-2025-08-07 &
  0.40 & \textbf{0.97} & 0.49 & 0.44 & 0.72 & 0.72 & 0.17 & 0.44 & \textbf{0.57} & 0.56 & 0.26 & 0.18 & 0.83 & 0.52 \\
&
  gpt-5-nano-2025-08-07 &
  0.50 & 0.92 & 0.38 & 0.16 & 0.64 & 0.68 & 0.21 & 0.32 & 0.47 & 0.61 & 0.24 & 0.23 & 0.79 & 0.47 \\
&
  gpt-4o-2024-11-20 &
  0.49 & 0.95 & 0.52 & 0.41 & 0.64 & 0.74 & 0.13 & 0.56 & 0.44 & 0.42 & 0.22 & 0.18 & 0.80 & 0.50 \\
&
  gpt-4o-mini-2024-07-18 &
  0.46 & 0.96 & 0.44 & 0.40 & 0.55 & 0.80 & 0.13 & 0.55 & 0.54 & 0.58 & 0.21 & 0.19 & 0.79 & 0.51 \\
&
  gpt-4.1-2025-04-14 &
  0.39 & 0.96 & 0.52 & 0.36 & 0.61 & 0.74 & 0.14 & 0.53 & 0.50 & \textbf{0.74} & 0.12 & 0.18 & 0.83 & 0.51 \\
&
  gpt-4.1-mini-2025-04-14 &
  0.54 & 0.92 & 0.47 & 0.24 & 0.62 & 0.61 & 0.10 & \textbf{0.60} & 0.38 & 0.63 & 0.20 & 0.16 & 0.76 & 0.48 \\
&
  gpt-4.1-nano-2025-04-14 &
  0.53 & 0.81 & 0.47 & 0.16 & 0.57 & 0.53 & 0.17 & 0.48 & 0.51 & 0.70 & \textbf{0.46} & 0.29 & 0.81 & 0.50 \\

\bottomrule
\end{tabular}}
\caption{Classification of dimensions of trust and distrust by zero-shot and few-shot models, and the scores used for our criteria under the Any Criteria. Dimensions: Ben: Benevolence, Com: Competence, Dec: Deception, Dis: Dishonesty, Fam: Familiarity, Inc: Incompetence, Int: Integrity, Mal: Malevolence, Opa: Opaqueness, Rel: Reliability, Tra: Transparency, Unf: Unfamiliarity, Unr: Unreliability.
}
\label{tab:any_task_style}
\end{table*}

\section{Demographic Summary}\label{sec:app-demog}
Table~\ref{tab:demographics} summarizes the demographics of annotators whose responses were retained.

\begin{table*}[t]
\centering
\begin{minipage}{\columnwidth}
\centering
\resizebox{0.7\columnwidth}{!}{%
\begin{tabular}{llrr}
\toprule
 & Question & \# & (\%) \\
\midrule
 & 18-24 & 25 & 10.9\% \\
    & 25-34 & 77 & 33.5\% \\
    & 35-44 & 47 & 20.4\% \\
\multirow{2}{*}{\rotatebox{90}{Age}}    & 45-54 & 45 & 19.6\% \\
    & 55-64 & 25 & 10.9\% \\
    & 65 or older & 10 & 4.3\% \\
    & Prefer not to say & 1 & 0.4\% \\
\midrule
 & Male & 84 & 36.5\% \\
       & Female & 143 & 62.2\% \\
\multirow{2}{*}{\rotatebox{90}{Gender}}       & Non-binary / third gender & 2 & 0.9\% \\
       & Other & 0 & 0.0\% \\
       & Prefer not to say & 1 & 0.4\% \\
\midrule
 & Asian & 4 & 1.7\% \\
                & Black or African American & 64 & 27.8\% \\
                & Hispanic or Latino & 10 & 4.3\% \\
\multirow{2}{*}{\rotatebox{90}{Race}}  & Native American or Alaska Native & 1 & 0.4\% \\
               & Native Hawaiian or Pacific Islander & 1 & 0.4\% \\
               & White & 146 & 63.5\% \\
               & Other & 3 & 1.3\% \\
               & Prefer not to say & 1 & 0.4\% \\
\hline
  & Some high school, no diploma & 3 & 1.3\% \\
            & High school diploma or GED & 14 & 6.1\% \\
            & Some college, no degree & 26 & 11.3\% \\
\multirow{2}{*}{\rotatebox{90}{Education}}          & Associate (2-year) degree & 15 & 6.5\% \\
  & Bachelor's (4-year) degree & 95 & 41.3\% \\
           & Master's degree & 66 & 28.7\% \\
          & Doctorate degree & 9 & 3.9\% \\
          & Prefer not to say & 2 & 0.9\% \\
\hline
 & Employed full-time & 133 & 57.8\% \\
                  & Employed part-time & 59 & 25.7\% \\
\multirow{2}{*}{\rotatebox{90}{Employment}}                  & Student & 6 & 2.6\% \\
                  & Disabled & 2 & 0.9\% \\
                  & Retired & 6 & 2.6\% \\
                  & Unemployed & 22 & 9.6\% \\
                  & Prefer not to say & 2 & 0.9\% \\
\hline

 & \$150,000 or more & 22 & 9.6\% \\
              & \$75,000 to \$99,999 & 42 & 18.3\% \\
\multirow{3}{*}{\rotatebox{90}{Income}}              & \$100,000 to \$149,999 & 49 & 21.3\% \\
              & \$50,000 to \$74,999 & 39 & 17.0\% \\
              & \$30,000 to \$49,999 & 33 & 14.3\% \\
              & Under \$30,000 & 39 & 17.0\% \\
              & Prefer not to say & 6 & 2.6\% \\

\bottomrule
\end{tabular}}
\end{minipage}
\begin{minipage}{0.9\columnwidth}
\centering
\resizebox{0.6\columnwidth}{!}{%
\begin{tabular}{llrr}
\toprule
 & Question & \# & (\%) \\
\midrule
 & Democratic Party & 92 & 40.0\% \\
\multirow{2}{*}{\rotatebox{90}{Political Party}}                & Republican Party & 87 & 37.8\% \\
                & Libertarian Party & 4 & 1.7\% \\
                & Green Party & 1 & 0.4\% \\
                & Independent & 36 & 15.7\% \\
                & Other & 3 & 1.3\% \\
                & Prefer not to say & 7 & 3.0\% \\

\hline
 & English & 229 & 99.6\% \\
\multirow{2}{*}{\rotatebox{90}{Native Language}}                & Spanish & 0 & 0\% \\
                & Chinese & 0 & 0\% \\
                & Hindi & 0 & 0\% \\
                & Arabic & 0 & 0\% \\
                & Other & 0 & 0\% \\
                & Prefer not to say & 1 & 0.4\% \\
\hline
 & Expert
 & 38 & 16.5\% \\
\multirow{2}{*}{\rotatebox{90}{Tech Level}}                  & Advanced
 & 108 & 47.0\% \\
                  & Intermediate
 & 78 & 33.9\% \\
                  & Beginner
 & 5 & 2.2\% \\
                  & Prefer not to say & 1 & 0.4\% \\
\hline
 & Not familiar
 & 7 & 3.0\% \\
 \multirow{1}{*}{\rotatebox{90}{LLM Familiar}}               & Somewhat familiar
 & 31 & 13.5\% \\
                & Moderately familiar
 & 83 & 36.1\% \\
                & Very familiar
 & 90 & 39.1\% \\
                & Expert
 & 18 & 7.8\% \\
                & Prefer not to say & 1 & 0.4\% \\
\hline
 & Never & 9 & 3.9\% \\
\multirow{1}{*}{\rotatebox{90}{LLM Use}}                   & Rarely & 29 & 12.6\% \\
                   & Occasionally & 56 & 24.3\% \\
                   & Frequently & 106 & 46.1\% \\
                   & Always & 29 & 12.6\% \\
                   & Prefer not to say & 1 & 0.4\% \\
\hline
 & Fully trust & 27 & 11.7\% \\
              & Somewhat trust & 134 & 58.3\% \\
\multirow{1}{*}{\rotatebox{90}{Trust LLMs}}              & Not sure & 32 & 13.9\% \\
            & Somewhat distrust & 32 & 13.9\% \\          
              & Fully distrust & 3 & 1.3\% \\    
              & Prefer not to say & 2 & 0.9\% \\
\bottomrule
\end{tabular}}
\end{minipage}
\caption{Summary of Annotators' Demographics}
\label{tab:demographics}
\end{table*}